\documentclass[numsec,webpdf,modern,large,namedate]{oup-authoring-template}

\onecolumn % for one column layouts

\graphicspath{{figures/}}

\usepackage{aliascnt}

\makeatletter
\newtheorem*{rep@theorem}{\rep@title}
\newcommand{\newreptheorem}[2]{%
\newenvironment{rep#1}[1]{%
 \def\rep@title{#2 \ref{##1}}%
 \begin{rep@theorem}}%
 {\end{rep@theorem}}}
\makeatother

\theoremstyle{thmstyleone}%
\newtheorem{theorem}{Theorem}%  meant for continuous numbers
\theoremstyle{thmstyletwo}%
\newtheorem{example}{Example}%
\newreptheorem{example}{Example}
\newtheorem{remark}{Remark}%

\theoremstyle{thmstylethree}%
\newtheorem{definition}{Definition}
\newtheorem{assumption}{Assumption}%

\AtEndDocument{\refstepcounter{theorem}\label{finalthm}}
\AtEndDocument{\refstepcounter{assumption}\label{finalasmp}}
\AtEndDocument{\refstepcounter{definition}\label{finaldef}}
\AtEndDocument{\refstepcounter{remark}\label{finalremark}}
\usepackage{balance}
\usepackage{wrapfig}
\usepackage{times}
\usepackage{mathtools}
\usepackage{colortbl}
\usepackage{bm, bbm}
\usepackage{cleveref}
\usepackage{booktabs}

\usepackage[shortlabels]{enumitem}
\usepackage[normalem]{ulem}
\usepackage{pifont}% http://ctan.org/pkg/pifont
\usepackage{xspace}
\usepackage{multirow}
\usepackage{diagbox}
\usepackage{caption}

\usepackage{eqparbox,array}

\algrenewcommand{\algorithmiccomment}[1]{\hfill\# {#1}}

\allowdisplaybreaks
\usepackage{hyperref}
\usepackage{xr}

\makeatletter
\newcommand\externaldocumentnocite[2][]{%
  \begingroup
  \let\bibcite\@gobbletwo   % ignore \bibcite{<key>}{...}
  \let\citation\@gobble     % ignore \citation{...}
  \let\bibdata\@gobble      % ignore \bibdata{...}
  \let\bibstyle\@gobble     % ignore \bibstyle{...}
  \externaldocument[#1]{#2} % read labels from #2.aux
  \endgroup
}
\makeatother

\makeatletter
\newcommand*{\addFileDependency}[1]{% argument=file name and extension
\typeout{(#1)}% latexmk will find this if $recorder=0
\@addtofilelist{#1}
\IfFileExists{#1}{}{\typeout{No file #1.}}
}
\newcommand*{\myexternaldocument}[1]{%
\externaldocumentnocite{#1}%
\addFileDependency{#1.tex}%
\addFileDependency{#1.aux}%
}

\myexternaldocument{jrssb_supp} % Overleaf cross reference

\newcommand{\Real}{\mathbb{R}}
\newcommand{\AIC}{\textrm{GIC}_{\lambda_n}}
\renewcommand{\hat}{\widehat}
\renewcommand{\tilde}{\widetilde}
\def\ind{\mathbbm{1}}
\def\E{\mathbb{E}} %expectation
\def\P{\mathbb{P}} %prob
\def\T {{ \mathrm{\scriptscriptstyle T} }} %transpose
\def\var{\textit{var}} %var
\def\cov{\textit{cov}}
\def\corr{\textit{corr}}

\newcommand{\fd}{f^*}
\newcommand{\ft}{f_{N}}
\newcommand{\ul}{u}
\newcommand{\Ul}{U}
\newcommand{\pl}{\textrm{PL}}
\newcommand{\dm}{M}
\newcommand{\dmc}{\mathcal{M}}
\newcommand{\qs}{Q}
\newcommand{\qsc}{\mathcal{Q}}
\newcommand{\la}{T}
\newcommand{\lac}{\mathcal{T}}

\newcommand{\Y}{\tilde{Y}}
\newcommand{\fc}{\mathcal{F}}
\newcommand{\X}{ X}
\def\x{x}
\bmdefine\w{w}
\bmdefine\y{y}
\bmdefine\z{z}
\bmdefine\e{e}
\bmdefine\u{u}
\bmdefine\bv{v}
\bmdefine\h{h}
\bmdefine\bz{0}
\bmdefine\balpha{\alpha}
\bmdefine\bbeta{\beta}
\bmdefine\bgamma{\gamma}
\bmdefine\bmu{\mu}
\def\loss{\ell}

\newcommand{\defeq}{:=}
\newcommand{\rv}[1]{{ #1}}
\def\dno{\dm_{\textrm{NO}}}
\def\drn{\dm_{\textrm{IID}}}
\def\dun{\dm_{\textrm{Const}}}
\def\dour{\dm_{\textrm{Poly}}}

\def\qiid{\qs_{\textrm{IID}}}

\DeclareMathOperator*{\dgeq}{\dot{\gtrsim}}

\newcommand\numberthis{\addtocounter{equation}{1}\tag{\theequation}}

\DeclareMathOperator*{\argmin}{\arg\min}

\DeclareMathOperator*{\argmax}{\arg\max}
\DeclarePairedDelimiterX{\norm}[1]{\lVert}{\rVert}{#1}
\DeclarePairedDelimiterX{\bnorm}[1]{\biggl\lVert}{\biggr\rVert}{#1}
\DeclarePairedDelimiterX{\abs}[1]{\lvert}{\rvert}{#1}
\DeclareMathOperator*{\esssup}{ess\,sup}
\DeclarePairedDelimiterX{\ip}[2]{\langle}{\rangle}{#1,#2} % inner product

\makeatletter
\newcommand\Autoref[1]{\@first@ref#1,@}
\def\@throw@dot#1.#2@{#1}% discard everything after the dot
\def\@set@refname#1{%    % set \@refname to autoefname+s using \getrefbykeydefault
    \edef\@tmp{\getrefbykeydefault{#1}{anchor}{}}%
    \xdef\@tmp{\expandafter\@throw@dot\@tmp.@}%
    \ltx@IfUndefined{\@tmp autorefnameplural}%
         {\def\@refname{\@nameuse{\@tmp autorefname}s}}%
         {\def\@refname{\@nameuse{\@tmp autorefnameplural}}}%
}
\def\@first@ref#1,#2{%
  \ifx#2@\autoref{#1}\let\@nextref\@gobble% only one ref, revert to normal \autoref
  \else%
    \@set@refname{#1}%  set \@refname to autoref name
    \@refname~\ref{#1}% add autoefname and first reference
    \let\@nextref\@next@ref% push processing to \@next@ref
  \fi%
  \@nextref#2%
}
\def\@next@ref#1,#2{%
   \ifx#2@ and~\ref{#1}\let\@nextref\@gobble% at end: print and+\ref and stop
   \else, \ref{#1}% print  ,+\ref and continue
   \fi%
   \@nextref#2%
}
\makeatother

\makeatletter
\let\oldtheequation\theequation
\renewcommand\tagform@[1]{\maketag@@@{\ignorespaces#1\unskip\@@italiccorr}}
\renewcommand\theequation{(\oldtheequation)}
\makeatother

\makeatletter
\providecommand{\listoffigures}{\section*{Figure Legends}\@starttoc{lof}}
\makeatother

\usepackage[doublespacing]{setspace}
\usepackage[fontsize=12pt]{fontsize}

\begin{document}

\journaltitle{Journals of the Royal Statistical Society}
\DOI{DOI HERE}
\copyrightyear{XXXX}
\pubyear{XXXX}
\access{Advance Access Publication Date: Day Month Year}
\appnotes{Original article}

\firstpage{1}

%\subtitle{Subject Section}

\title[Understand Model Stealing Attack and Defense]{Model Privacy: A Unified Framework for Understanding Model Stealing Attacks and Defenses}

\author[1]{Ganghua Wang}
\author[1]{Yuhong Yang}
\author[1,$\ast$]{Jie Ding} %\ORCID{0000-0000-0000-0000}

\authormark{Author Name et al.}

\address[1]{\orgdiv{School of Statistics}, \orgname{University of Minnesota}, \orgaddress{\street{224 Church Street}, \postcode{55455}, \state{Minnesota}, \country{USA}}}

\corresp[$\ast$]{Corresponding author. \href{dingj@umn.edu}{dingj@umn.edu}}

\received{Date}{0}{Year}
\revised{Date}{0}{Year}
\accepted{Date}{0}{Year}

\abstract{
    The use of machine learning (ML) has become increasingly prevalent in various domains, highlighting the importance of understanding and ensuring its safety. One pressing concern is the vulnerability of ML applications to model stealing attacks. These attacks involve adversaries attempting to recover a learned model through limited query-response interactions, such as those found in cloud-based services or on-chip artificial intelligence interfaces. While existing literature proposes various attack and defense strategies, these often lack a theoretical foundation and standardized evaluation criteria. In response, this work presents a framework called ``Model Privacy'', providing a foundation for comprehensively analyzing model stealing attacks and defenses. We establish a rigorous formulation for the threat model and objectives, propose methods to quantify the goodness of attack and defense strategies, and analyze the fundamental tradeoffs between utility and privacy in ML models. Our developed theory offers valuable insights into enhancing the security of ML models, especially highlighting the importance of the query-dependent structure of perturbations for effective defenses. We demonstrate the application of model privacy from the defender's perspective through various learning scenarios.  Extensive experiments corroborate the insights and the effectiveness of defense mechanisms developed under the proposed framework.
}
%% SUBMITTED VERSION ON sept 19
% The use of machine learning (ML) has become increasingly prevalent in various domains, highlighting the importance of understanding and ensuring its safety. One pressing concern is the vulnerability of ML applications to model stealing attacks. These attacks involve adversaries attempting to recover a learned model through limited query-response interactions, such as those found in cloud-based services or on-chip artificial intelligence interfaces. While existing literature proposes various attack and defense strategies, these often lack a theoretical foundation and standardized evaluation criteria. In response, this work presents a framework called ``Model Privacy'', providing a foundation for comprehensively analyzing model stealing attacks and defenses. We establish a rigorous formulation for the threat model and objectives, propose methods to quantify the goodness of attack and defense strategies, and analyze the fundamental tradeoffs between utility and privacy in ML models. Our developed theory offers valuable insights into enhancing the security of ML models, especially highlighting the importance of the attack-specific structure of perturbations for effective defenses. We demonstrate the application of model privacy from the defender's perspective through various learning scenarios.  Extensive experiments corroborate the insights and the effectiveness of defense mechanisms developed under the proposed framework.


% \boxedtext{
% \begin{itemize}
% \item Key boxed text here.
% \item Key boxed text here.
% \item Key boxed text here.
% \end{itemize}}

\maketitle

\section{Introduction} \label{sec:intro}
    
    Machine learning (ML) has achieved remarkable success in many applications such as content recommendation on social media platforms, medical image diagnosis, and autonomous driving. However, there is a growing concern regarding the potential safety hazards coming with ML. One particularly critical threat to ML applications is model stealing attacks~\citep{tramer2016stealing, oliynyk2023know}, where adversaries attempt to recover the learned model itself through limited query-response interactions. ML models are often highly valuable due to their expensive creation process and proprietary techniques. If a third party can reconstruct the model, it may significantly harm the model owner's interests, leading to financial losses and intellectual property breaches. Model stealing attacks target trained models. This differs from attacks aiming to compromise datasets, which are often studied under the framework of differential privacy or its variants~\citep[see, e.g.,][]{evfimievski2003limiting, dwork2006calibrating, dong2022gaussian}, \rv{as further elaborated in Section 1 of the supplementary document. }

    Particularly, Machine-learning-as-a-service, a concept that gained increasing popularity since the development of LLMs~\citep{philipp2020machine, gan2023model}, is a natural target of model stealing attacks. 
    For instance, suppose a model owner trains a classification model and aims to monetize it by deploying it on a cloud server and providing paid service. Users can send queries, which in this case are images, to the server and receive model-generated labels as responses after paying for the service. However, model stealing attacks pose a serious threat to this monetization strategy, since a malicious user may easily rebuild a model with comparable accuracy to the original model based on a few query-response pairs. For instance, \citet{tramer2016stealing} reported that 100 queries are sufficient for a successful model stealing attack on a classification model trained on the handwritten digits dataset, costing less than one dollar. 
    
    Empirical studies have shown that model stealing attacks can be very effective across a wide range of applications, such as classification tasks with logistic regression, random forests, and deep neural networks~\citep{tramer2016stealing,papernot2016practical, orekondy2019knockoff}. Our experiments further demonstrate that for large language model-based classifiers, such as those based on GPT~\citep{radford2018improving}, BERT~\citep{devlin2018bert}, and XLNet~\citep{yang2019xlnet}, one could rebuild a model with similar predictive accuracy by querying only 5\% of the original model's training sample size, if there is no defense. The consequences of such attacks are significant: the attacker can stop paying for the service, potentially leading to financial losses for the model owner, or even compete with the original model owner using the stolen model. These risks highlight the urgent need for effective defense mechanisms to protect the integrity and value of ML models.

    % \begin{figure}
    %     \centering
    %     \includegraphics[width=0.8\linewidth]{figures/FigIntro.pdf}
    %     \caption{An illustration of a model stealing attack in the machine-learning-as-a-service scenario. Given a trained image classification model deployed on the cloud server, a user can send queries to and receive responses from the server. A malicious user, also called an attacker, aims to reconstruct this model based on query-response pairs. \gw{Referee 1 suggested move it to Appendix?}}
    %     \label{fig:MLaaS}
    % \end{figure}

    A large body of literature has been developed for both model stealing attacks and defenses~\citep[see, e.g.,][more details in Section~\ref{sec:supp_related} of the supplementary document.]{tramer2016stealing,papernot2016practical,chandrasekaran2018model,milli2019model,kesarwani2018model,juuti2019prada,lee2019defending,orekondy2019prediction,wang2020information,oliynyk2023know}. However, existing studies are often heuristic and based on practical experience, lacking a standard criterion for evaluating the effectiveness of attack and defense methods. This makes it challenging to compare different approaches. Moreover, the principles guiding which defense a service provider should adopt remain unclear. Establishing a methodological foundation for this subject is therefore crucial, and this paper aims to address this gap.

    We create a statistical framework named \textit{model privacy} to serve as a foundation for analyzing model stealing attacks and defenses. Our contributions are both conceptual and theoretical. \textbf{First,} we establish a formal conceptual foundation for model stealing attacks. In particular, we identify fundamental factors such as attacker's query and learning strategy and defender's defense mechanism, quantify the effectiveness of model stealing attacks and defenses through statistical risk, and formalize the actions and objectives of both attackers and defenders. 
    \textbf{Second,} we theoretically analyze how to choose optimal defense mechanisms from the defender's perspective. Under the model privacy framework, we evaluate the privacy-utility tradeoffs of multiple defense mechanisms against several representative learning algorithms ($k$-nearest neighbor, polynomial regression, kernel ridge regression, and neural networks) and assess their worst-case privacy guarantees against unknown attackers. New defense mechanisms are proposed for both regression and classification tasks with theoretical guarantees. 
    Our analysis yields two intriguing findings: (1) An attacker can easily reconstruct the target model when the defender does not defend or merely adds IID noise, highlighting the need for more sophisticated defense mechanisms; and (2) Crafting attack-specific, query-dependent perturbations is critical for effective defenses. 
    For instance, in the case of polynomial regression, while an attacker can exactly recover a model after only finitely many queries if there is no defense, under our proposed defense mechanism, Order Disguise, the attacker cannot build a model that closely approximates the defender’s.  
    Numerical results on both simulated and real-world datasets show that our proposed defenses significantly enhance model privacy compared to baseline methods such as adding IID noise.
   
    The rest of this paper is organized as follows. \Autoref{sec:motivation} formulates the model privacy framework, including threat model, goodness quantification, and objectives. It also provides insights into the proposed concepts. 
    % \Autoref{sec:formulation} proposes practical measures for evaluating the goodness of attacks and defenses, and establishes the connection with the rebuilt model's statistical risk. 
    % \Autoref{subsec:notation} introduces the setup and notations used throughout the paper. 
    In \Autoref{sec:model_func}, we apply the model privacy framework to understand the defender's optimal decisions in various scenarios where the attacker's learning algorithm is known. \Autoref{sec:atk} focuses on defense analysis when the attacker's learning algorithm is unknown to the defender. Numerical experiments on both simulated and real-world datasets are presented in \Autoref{sec:exp} to corroborate our findings. We conclude the paper with further discussions in \Autoref{sec:con}. The supplementary document includes proofs and additional theoretical and experimental results.

% \section{Related Work} \label{subsec:pastwork}
    % \input{relatedwork}

\section{Formulation of the Model Privacy Framework} \label{sec:motivation}
    \def\attstr{\textrm{AttStr}}
\def\atteff{\textrm{AttEff}}
\def\relatteff{\textrm{RelAttEff}}
\def\defstr{\textrm{DefStr}}
\def\defeff{\textrm{DefEff}}
\def\reldefeff{\textrm{RelDefEff}}
\def\utilloss{\textrm{UtilLoss}}
\def\eco{potent\xspace}

    In \Autoref{subsec:overview}, we present the setup of model stealing attacks, highlighting the motivation and essential ingredients of the proposed framework. 
    In \Autoref{subsec:quantify}, we propose methods to quantify the effectiveness of attack and defense strategies.
    In \Autoref{subsec:game}, we characterize the objectives and actions of both the attacker and defender, providing a foundation for theoretically analyzing the unique challenges in model privacy. 
    Additionally, an economic perspective of model privacy and a generalized formulation of objectives are presented in Sections~\ref{sec:supp_eco} and~\ref{sec:supp_game} of the supplementary document.

     \newcommand{\hld}{H}
        \textbf{Notations.} 
        For two non-negative sequences $\{a_n, n=1,2,\ldots\}$ and $\{b_n, n=1,2,\ldots\}$, $a_n \lesssim b_n$ means that $\limsup_{n\to \infty} a_n/b_n < \infty$, also denoted as $a_n = O(b_n)$; $a_n \gtrsim b_n$ means $b_n \lesssim a_n$; $a_n \asymp b_n$ means both $a_n \lesssim b_n$ and $a_n \gtrsim b_n$; and $a_n=o(b_n)$ means $\lim_{n\to \infty} a_n/b_n=0$. A dot over $\lesssim, \gtrsim, \asymp$ means up to a polynomial of $\ln(n)$. We denote the expectation, probability, covariance, correlation, and indicator function as $\E$, $\P$, $\cov$, $\corr$, and $\ind_{(\cdot)}$, respectively. For a vector $\w=(w_1, \ldots, w_d)^\T \in \Real^d$, its $\ell_p$-norm ($p \geq 0$) is $\norm{\w}_0 = \sum_{i=1}^d \ind_{w_i=0}, \ 
            \norm{\w}_p = \biggl\{\sum_{i=1}^d \abs{w_i}^p\biggr\}^{1/p} \textrm{ for } p>0.$
        % \begin{align*}
        %     \norm{\w}_0 = \sum_{i=1}^d \ind_{w_i=0}, \quad 
        %     \norm{\w}_p = \biggl\{\sum_{i=1}^d \abs{w_i}^p\biggr\}^{1/p} \textrm{ for } p>0.
        % \end{align*}
        % The $k$-th derivative of $f$ is denoted as $f^{(k)}$.
        For $q > 0$, the $L_q(\mu)$-norm of $f$ is $\norm{f}_\infty = \esssup_{\x} \abs{f(\x)}, \  
 \norm{f}_q = \biggl(\int |f(\x)|^q\mu(d\x)\biggr)^{1/q}.$
 %        \begin{align*}
 %             \norm{f}_\infty = \esssup_{\x} \abs{f(\x)}, \quad 
 % \norm{f}_q = \biggl(\int |f(\x)|^q\mu(d\x)\biggr)^{1/q}\textrm{ for } q>0 .
 %        \end{align*}  
        For ease of reference, we have summarized the key notations (some introduced later) used throughout the paper in \Autoref{tab:notation_1}. 

        \begin{table}[tb]
            \centering
            \begin{tabular}{lp{13.5cm}}
                \toprule
                \textbf{Symbol} & \textbf{Meaning} \\
                 \midrule
                $n$, $d$ & Sample size (total number of queries sent by the attacker), dimension of input $\X$\\ \midrule
                $\fd$, $\hat{f}_n$, $\fc$ & Defender's model, attacker's reconstructed model, and defender's function class\\ \midrule
                $\dm, \ul_n(\dm)$ & Defense mechanism and its utility loss\\\midrule
                $\qs, \la$ & Attacker's query strategy and learning algorithm \\
                % $\dm, \dmc$ & Defense mechanism, and its possible collection \\
                % \midrule
                % $\la, \lac$ & Attacker's learning algorithm, and its possible collection \\
                % \midrule
                % $\qs, \qsc$ & Attacker's query strategy, and its possible collection \\
                % \midrule
                % $\si$ & Attacker's and defender's side information \\
                % \midrule
                % $\ell_A, \ell_D$ & Attacker's and defender's loss function \\
                \midrule
                $\X_i$, $\Y_i$, $e_i$ & A particular query, its response, and corresponding perturbation\\ \midrule
                % $Z_n$ & The query-response pairs $\{(\X_i, \Y_i),i=1,\dots,n\}$\\ \midrule
                $\pl_n(\dm,\fc \mid \qsc, \lac)$ & Privacy level of a defense mechanism against attack strategies $(\qsc, \lac)$ \\ 
                % \midrule
               % $r_n(\qs, \la;\mid \fc,\dmc)$ & Stealing error of an attack strategy given defense mechanisms $\dmc$ \\ 
               % \midrule
                % $\ul_n(\dm)$ & Utility loss of a defense mechanism \\ \midrule
                % $\Ul_n$ & A given utility loss level  \\ 
                 \bottomrule
            \end{tabular}
            \caption{A summary table of frequently used notations.}
            \label{tab:notation_1}
        \end{table}

    \subsection{Background and Essential Ingredients of Model Privacy} \label{subsec:overview}
    We begin with a classical machine learning setting. Suppose a model owner has created a function $\fd$ from a large proprietary dataset to predict responses based on input $\X$. The model owner provides query-based services by responding with $\Y_i$ for any input query $\X_i$. 
    
    Meanwhile, a malicious user, or attacker, aims to build a model to either provide similar services and compete with the model owner or to discontinue paying for the service. Instead of collecting raw training data and training a model from scratch, the attacker exploits the model owner's service to significantly reduce training costs. In particular, the attacker performs a model stealing attack to reconstruct a function $\hat{f}_n$ that closely approximates $\fd$ using $n$ query-response pairs $Z_n\defeq \{(\X_i, \Y_i),i=1,\dots,n\}$. There are three essential ingredients for the attacker.
    \begin{enumerate}
        \item \textit{ Evaluation criterion} $\ell$. To evaluate the goodness of an attack, the attacker has to assess the closeness between the target function $\fd$ and the reconstructed function $\hat{f}_n$. This is measured via $\ell_A(\hat{f}_n, \fd)$, where $\ell_A$ is a loss function chosen according to the attacker's interest. For example, if the attacker wants $\hat{f}_n$ to be uniformly close to $\fd$, $\ell_A(f, g)$ could be the supremum norm $\sup_x \abs{f(x)-g(x)}$ for two functions $f$ and $g$. Alternatively, if the focus is on performance in a specific region $S$, $\ell_A(f, g)$ can be defined as $\int_{x \in S} \abs{f(x) - g(x)} dx$.
        
        \item \textit{ Attack strategy} $(\qs, \la)$. An attack strategy consists of a \textit{query strategy} $\qs$ and a \textit{learning algorithm} $\la$. 
        A learning algorithm $T$ is used by the attacker to reconstruct or sequentially update models based on the observed query-response pairs.
        Common learning algorithms include e.g., linear regression, $k$-nearest neighbor, ensemble forests, and neural networks. 
        The query strategy $\qs$ determines how the attacker selects $\X_i$'s. There are two types of query strategies: ``batch query'', where all $n$ queries are sent to the model owner at once, and ``sequential query'', where queries are sent one by one or in parts.
        Examples of batch query strategies include fixed designs (e.g., equally-spaced sampling over an interval) and random designs (e.g., IID sampling from a given distribution). Sequential query strategies may use previous responses or models trained by $T$ to guide an active choice of the subsequent $\X_i$.

        \item \textit{ Side information}.  
        Attacker's side information refers to any additional information available to the attacker beyond the query-response pairs. This may include properties of the defender's model, such as knowing that $\fd$ is monotone or is a neural network with a particular architecture, or additional data collected from other sources. Side information can enhance the attacker's learning capability.
    \end{enumerate}

    % Overall, the attacker aims to choose a favorable attack strategy that allows them to accurately recover $\fd$ using a minimal number of queries.

    Given the threat of stealing, the model owner aims to prevent the attacker from easily stealing the model. As such, we name this problem as enhancing \textit{model privacy} and call the model owner the defender. There are also three key components for the defender.
    \begin{enumerate}
        \item \textit{ Evaluation criterion} $\ell$. The defender seeks to maximize the dissimilarity between $\hat{f}_n$ and $\fd$, evaluated through $\ell_D(\hat{f}_n, \fd)$. It should be noted that the defender's loss function $\ell_D$ may differ from the attacker's $\ell_A$. For example, an attacker who aims to steal a language model for a given downstream task may employ a loss function $\ell_A$ evaluated in the task-specific region $S$. Meanwhile, the defender might aim for a comprehensive defense, therefore using a different $\ell_D$ that evaluates the entire input space.

        \item \textit{ Defense mechanism} $\dm$. 
        The defender can enhance model privacy by adding perturbations to the responses, $\Y_i=\fd(\X_i)+e_i$, where $e_i$ is a perturbation determined by a defense mechanism $\dm$. Larger perturbations can increase the deviation between $\hat{f}_n$ and $\fd$. However, they also reduce the quality of service for benign users, which we refer to as \textit{utility loss}. Every defense mechanism must navigate this trade-off between privacy and utility, which is referred to as the privacy-utility trade-off.

        \item \textit{ Side information}. The defender's side information refers to any knowledge about the attacker beyond the received queries. Examples include knowledge of the attacker's potential query strategies and learning algorithms. This information helps the defender deploy more targeted defense mechanisms, thereby bolstering model protection.
    \end{enumerate}

    \subsection{Goodness Quantification via Statistical Risk}\label{subsec:quantify}
        Among key ingredients identified in \Autoref{subsec:overview}, the attack strategy and defense mechanism are of most importance. 
        In this section, we demonstrate that understanding the stealing capability, characterized by the statistical risk of the rebuilt function $\E\{ \ell(\hat{f}_n, \fd)\}$, is essential for evaluating the goodness of defenses and attacks. 
        The following two examples shed lights on this insight, with $\ell$ being the squared error loss.

        \begin{example}[Stealing in a Parametric Setting]\label{ex:lin}
        Suppose the defender has a linear function $\fd$ of dimension $d$, and the attacker tries to rebuild it by fitting a linear model. Without any defense, the attacker can precisely reconstruct $\fd$ with as few as $d$ query-response pairs by solving a linear equation. 
        In contrast, when the defender adds IID noise to its responses, the attacker needs $n \asymp \epsilon^{-1}$ data points to obtain a model $\hat{f}_n$ such that $\E \norm{\hat{f}_n-\fd}_2^2 \lesssim \epsilon$, which can be significantly larger than $d$ for a small $\epsilon$, incurring a higher cost for the attacker due to more queries needed. 
        % Without querying the defender, the attacker can also collect IID data from nature, which will be called nature data, and then build a model. For nature data with noise variance $\sigma^2 > 0$, the attacker needs $n \asymp \epsilon^{-1}$ data points to obtain a model $\hat{f}_n$ such that $\norm{\hat{f}_n-\fd}_2^2 \lesssim \epsilon$, which can be significantly larger than $d$, incurring a high data collection cost. 
        \end{example}

        \begin{example}[Stealing in a Non-Parametric Setting]\label{ex:steal_nonpara}
            Now, suppose the attacker uses a non-parametric learning algorithm, such as $k$-nearest neighbor. The convergence rate in this case is known for a variety of familiar regression function classes. Taking the Lipschitz function class on $[0,1]^d$ for example, we typically have $\E \norm{\hat{f}_n- \fd}_2^2 \asymp n^{-2/(d+2)}$ when IID noise with a constant order variance is injected into the responses. 
            % Therefore, the necessary stealing sample size is at the order of $\epsilon^{-(d+2)/2}$.
            If the defender adds long-range dependent noise into the responses, we will prove in \Autoref{sec:atk} that $\E \norm{\hat{f}_n- \fd}_2^2 \asymp n^{-\gamma}$, where $\gamma$ is a small number in $(0, 1)$ that represents the degree of noise correlation. 
            Consequently, it is more difficult for the attacker to recover $\fd$ with responses perturbed by long-range dependent noise.
            % Consequently, the necessary stealing sample size $n(\epsilon,\ell;\fd,\dm,\qs,\la) \asymp \epsilon^{-1/\gamma}$, implying that adding long-range dependent noise is a \eco defense when $\gamma < 2/(d+2)$. 
        \end{example}
       
        % Practitioners may consider the goodness of attack and defense with additional perspectives. Regardless, a
        As seen from the examples above, the essential issue in goodness quantification is \textit{how well an attacker can learn the underlying function with $n$ queries}, which we call the stealing capability. The rebuilt model's statistical risk $\E \{ \ell(\hat{f}_n, \fd)\}$ captures the similarity between $\hat{f}_n$ and $\fd$, hence can be used to characterize the attack and defense goodness.
        Based on the worst-case convergence rate over a function class $\fc$ that contains $\fd$, we define the following privacy level to measure the goodness of a defense $\dm$ against a collection of possible attack strategies. Analogously, we define the stealing error to measures the goodness of an attack $(\qs,\la)$ against a collection of possible defenses. 

        \begin{definition}[Privacy Level of a Defense]\label{def:priv}
            The (worst-case) privacy level of a defense mechanism $\dm$ against a collection of query strategies $\qsc$ and learning algorithms $\lac$ at a sample size $n$ is 
             \begin{align*}
                \pl_n(\dm,\fc \mid \qsc,\lac) \defeq  \inf_{\qs \in \qsc,\la \in \lac} \sup_{\fd \in \fc} \E \{ \ell_D(\fd, \hat{f}_n) \}. 
            \end{align*}
        Here, $\hat{f}_n$ depends on $\qs$, $\la$, $\fd$, and $\dm$, since it is reconstructed from the query-response pairs.
        % A defense mechanism $\dm^*$ is called rate optimal within a collection of defenses $\dmc$ if
        % \begin{align*}
        %     \pl_n(\dm^*,\fc \mid \qsc,\lac) \asymp \sup_{\dm \in \dmc}  \pl_n(\dm,\fc \mid \qsc,\lac).
        % \end{align*}
        \end{definition}

        % Similarly, facing a collection of possible defenses $\dmc$, the attacker can evaluate the goodness of their attack strategy via the worst-case statistical risk as well.

        \begin{definition}[Stealing Error of an Attack Strategy]\label{def:stealerror}
            The (worst-case) stealing error of an attack strategy $(\qs,\la)$ with respect to a collection of possible defenses $\dmc$ at sample size $n$ is defined as 
            \begin{align*}
                r_n(\qs, \la \mid \fc,\dmc) \defeq \sup_{\fd \in \fc, \dm \in \dmc} \E \{ \ell_A(\fd, \hat{f}_n) \}.
            \end{align*}
        % An attack strategy $(\qs^*, \la^*)$ is called rate optimal within $(\qsc,\lac)$ if $ r_n(\qs^*, \la^* \mid \fc,\dmc) \asymp \inf_{\qs \in \qsc, \la \in \lac}  r_n(\qs, \la \mid \fc,\dmc) $.
        \end{definition}

        Regarding interpretation, the privacy level provides a guarantee of a defense mechanism's performance against a collection of potential attack strategies. For instance, if the attacker uses $k$-nearest neighbor algorithms with $k\geq 1$ to steal a Lipschitz function on $[0, 1]^d$, a defender injecting IID noise with a fixed noise variance achieves a privacy level at the order of $n^{-2/(d+2)}$, as will be seen in \Autoref{thm:knn}. Consequently, the defender can expect that the risk of the attacker's reconstructed function is at least at the order of $n^{-2/(d+2)}$, even under the best choice of $k$. Analogously, the stealing error provides a guarantee on $\hat{f}_n$'s performance from the attacker's perspective, thus quantifying the effectiveness of an attack strategy. 
        % the worst-case scenario for a defender deploying $\dun$ corresponds to an attacker using $k$-NN with $k = 1$.

        The privacy level is closely related to the stealing error when the defender and attacker have the same evaluation criteria. 
        % Specifically, given a function class $\fc$ that $\fd$ can be any function in it, the privacy level of a defense mechanism $\dm$ against $(\qsc,\lac)$ characterizes the smallest stealing error that an attacker can uniformly achieve by choosing an attack strategy in $(\qsc,\lac)$. 
         % and $\pl_n(\dm^*,\fc \mid \qsc, \lac) \lesssim  r_n(\qs^*, \la^* \mid \fc,\dmc)$
        In fact, for any given defense $\dm$ and attack $(\qs,\la)$, we have $\pl_n(\dm,\fc \mid \qs,\la) =  r_n(\qs, \la \mid \fc,\dm)$ when $\loss_A=\loss_D$, since both privacy level and stealing error aim to describe the attacker's stealing capability though from competing perspectives. 
        Those two quantities can be less aligned when $\loss_A$ and $\loss_D$ are different. For instance, if $\loss_A$ and $\loss_D$ focus on distinct regions, it is possible that the privacy level is high while the stealing error is small. 
        Further discussion is included in Section~\ref{subsec:manipulate} of the supplementary document. 
        % \Autoref{subsec:manipulate}. 

        % \rv{
        % \begin{example}[Stealing a Linear Function]\label{ex:lin}
        %     Suppose the defender has a linear function $\fd$ of $d$ dimensions, and the attacker tries to rebuild it by fitting a linear model. Without any defense mechanism, the attacker can precisely reconstruct $\fd$ with as few as $d$ query-response pairs by solving a linear equation. 
        %     Without querying the defender, the attacker can also collect IID data from nature, which will be called nature data, and then build a model. 
        %     For nature data with noise variance $\sigma^2 > 0$,
        %     the attacker needs $n \asymp \epsilon^{-1}$ data points to obtain a model $\hat{f}_n$ such that $\norm{\hat{f}_n-\fd}_2^2 \lesssim \epsilon$, which can be significantly larger than $d$, incurring a high data collection cost. 
        % \end{example}
        % }

    It is important to keep in mind that privacy level itself is insufficient to compare the goodness of different defense mechanisms, because the defender has to consider both the service quality for the benign users and the difficulty for the attacker to reconstruct $\fd$ well.  
    In one extreme, suppose a defender deploys a defense of returning pure white noise. Clearly, the attacker cannot rebuild any useful model based on the defender's responses, but such a defense results in unacceptable service quality for benign users. 
    We thus define the utility loss of a defense $\dm$ to quantify the accuracy degradation from a benign user's perspective as follows.
    \begin{definition}[Utility Loss]\label{def:util}
            Let $X_1,X_2,\dots$ be queries sent from a benign user and $\loss_U$ be a loss function representing the benign user's interest. 
            The utility loss of a defense mechanism $\dm$ for this benign user at sample size~$n$ is defined as
            \begin{align*}
                \ul_n(\dm) \defeq \E \biggl\{ \frac{1}{n} \sum_{i=1}^n \loss_U(\fd(\X_i), \Y_i) \biggr\},
            \end{align*}
            where $\Y_i$ is the response perturbed by $\dm$. 
    \end{definition}       
    
    % \begin{definition}[Defense Strength]\label{def:defstr}
    %     The defense strength of a defense mechanism $\dm$ with respect to $\fd$ and an attack $(\qs, \la)$ is  
    %     \begin{align*}
    %     \defstr(\dm\mid \fd,\qs,\la,\ell_D) 
    %              \defeq   \liminf_{\epsilon \to 0^+} \frac{ n(\epsilon,\ell_D;\fd,\dm,\qs,\la) \times c_{\textrm{Steal}} }{n(\epsilon,\ell_D;\fd,\dnt,\qs,\la) \times c_{\textrm{Nature}} }.
    %     \end{align*}
    %     A defense mechanism $\dm$ with utility loss $\ul_n(\dm)$ is said to be \eco if $\defstr = \infty$.
    % \end{definition}

    Utility loss can be understood as the average difference between the perturbed response $\Y_i$ and the noiseless response $\fd(\X_i)$. 
    % While utility loss seems to be associated with the sample size $n$, 
    The defender can easily control utility loss at any desired level, e.g., by scaling the magnitude of the injected perturbations. 
    The defender therefore aims to achieve a high privacy level with a relatively small utility loss. This situation is similar to hypothesis testing, where the goal is to find a test having high power (true positive rate) and low significance level (false positive rate). Analogous to the concept of uniformly most powerful test \citep{casella2024statistical}, we propose to only compare defense mechanisms within a given level of utility loss, and a defense with a larger privacy level is considered better.
    % We therefore propose to only compare defense mechanisms with the same level of utility loss, and a defense with a larger privacy level is considered better.

    \subsection{Threat Model and Objectives of Model Privacy}\label{subsec:game}
        Building upon the essential elements and goodness quantification approach presented in previous subsections, this subsection formally states the threat model and the objectives of both the attacker and defender, completing our model privacy framework.

        \textbf{Threat Model.}
        We consider the following attacker-defender interactions. 
        The attacker sends $n$ queries $\X_1, \ldots, \X_n$ to the defender according to a query strategy $\qs$, possibly in a sequential manner. Upon receiving the queries, the defender determines the perturbations $e_i=\dm(\X_1, \ldots, \X_i), i=1,\dots,n$ for sequential queries or $(e_1, \dots, e_n)=\dm(\X_1, \ldots, \X_n)$ for batch queries, then returns responses $\Y_i = \fd(\X_i)+e_i$ to the attacker. Here, $\dm$ is a defense mechanism, and $\fd: \mathcal{X}\to \mathcal{Y}$ is the defender's trained function that needs protection. For convenience, we assume $\mathcal{X} \in \Real^d$ and $\mathcal{Y} \in \Real$ throughout this paper, though the proposed definitions and concepts can be extended to general input and response spaces.
        Given query-response pairs $Z_n = \{(\X_i, \Y_i), i=1,\ldots,n\}$, the attacker uses a learning algorithm $\la$ to reconstruct a model $\hat{f}_n=\la(Z_n)$. Specifically, $\la$ is a measurable mapping from $Z_n$ to a measurable function of $X$.

        % \begin{table}
        %     \centering
        %     \begin{tabular}{p{0.5in}p{0.5in}p{1.1in}p{3.2in}}
        %     \toprule
        %     Type & Notion & Name & Meaning \\ 
        %     \midrule
        %      \multirow{2}{*}{Query} & $\qiid$ & IID batch query & Draws an IID sample from a given distribution \\
        %       & $\qs_{\text{ES}}$   & Equally-spaced batch query  & Draws equally-spaced points in a given area \\ \hline
        %     \multirow{3}{*}{Defense}  & $\dno$ & No defense & $e_i=0$ for $i=1,\dots,n$ \\ 
        %       &$\drn$ & IID noising & $e_i$'s are IID drawn from a distribution with zero mean \\ 
        %       &$\dun$ & Constant noising & $e_i$ is a fixed constant for $i=1,\dots,n$ \\ 
        %       &$\dnt$ & Nature noising & $e_i$'s are drawn from the natural data-generating distribution \\ 
        %         \bottomrule
        %     \end{tabular}
        %     \caption{Summary of common attack and defense strategies.}
        %     \label{tab:common_strategy}
        % \end{table}

    \textbf{Objectives.}
    Both the attacker and defender aim to make the most advantageous decisions to maximize their interests. Specifically, given $\fc$ and $\dmc$, an attack strategy $(\qs^*, \la^*)$ is called the optimal among a collection of attacks $(\qsc, \lac)$ if
    \begin{align*}
            \qs^*, \la^* \in \argmin_{\qs \in \qsc, \la \in \lac} r_n(\qs, \la \mid \fc,\dmc). 
            \numberthis \label{eq:best_atk}
    \end{align*}
    Analogously, with respect to potential attack strategies $(\qsc, \lac)$, a defense mechanism $\dm^*$ is the optimal among a collection of defenses $\dmc$ if
        \begin{align*}
            \dm^{*} \in \argmax_{\dm \in \dmc}  \pl_n(\dm,\fc \mid \qsc,\lac), \  s.t.\  \ul_n(\dm) \leq \Ul_n. 
            \numberthis \label{eq:best_def}
        \end{align*}

            In efforts to find the optimal strategy for their respective objective, both the attacker and defender need to understand several fundamental issues in model privacy.
        Given the urgent societal concern of defending against model stealing attacks in real-world AI applications, this paper focuses primarily on the defender's perspective. We list some of the key problems below: 
        (1) When is it necessary to defend a model?
        (2) What is the best defense strategy?
        (3) What kinds of attacks are easier to defend against?
        (4) How do key ingredients listed in \Autoref{subsec:overview} affect decision-making? 
        Addressing these problems will provide theory-guided solutions for developing more effective defense strategies to secure ML model services. These issues are further discussed in \Autoref{sec:model_func, sec:atk}. In particular, \Autoref{sec:model_func} discusses the defender's decision-making given a known attack strategy, while \Autoref{sec:atk} considers defending against an attacker without any knowledge of their attack strategy.

    In summary, model privacy involves a non-cooperative and asymmetric interaction between an attacker and a defender.
    We establish the model privacy framework by defining key elements such as attack strategies, defense mechanisms, and their effectiveness, and formulating the threat model and objectives. This framework enables 
    us to systematically analyze and develop strategies to enhance model privacy, as studied in the subsequent sections. 

    \rv{\begin{remark}[Comparison with standard learning framework]
        The key distinction between model privacy and classical learning framework lies in the dependence structure of the observations. In model privacy, the responses are $\tilde{Y_i}=\fd(X_i)+e_i$, where the perturbations $e_i$ are generated according to a defense mechanism $\dm$ and can \textit{exhibit arbitrary dependence}. This setting fundamentally departs from the classical learning theory, which typically assumes that $e_i$'s are IID or exhibit structured dependence (e.g., short-range autoregressive dependence).
        In short, model privacy generalizes the standard learning scenario to the learning from arbitrarily dependent observations under a utility loss constraint. This generalization requires new tools to analyze learning from dependent observations.
    \end{remark}}

% \section{Analysis via Statistical Risk} \label{sec:formulation}
    % \input{formulation.tex}

\section{Defending against a Known Attack Learning Algorithm} \label{sec:model_func}

    This section examines the goodness of defense strategies when the defender knows that the attacker adopts a specific learning algorithm including $k$-nearest neighbor ($k$-NN) and polynomial regression. More learning scenarios are considered in Section~\ref{sec:supp_more_attacks} of the supplementary document, including kernel ridge regression, neural networks, and empirical risk minimization over general bounded function classes. In the supplement Section~\ref{subsec:manipulate}, we also explore the conditions under which defending is easier.
    Our findings help defenders assess the merits of different defense strategies and selecting the most suitable one. 
    Recall that each defense mechanism $\dm$ in comparison satisfies that $\ul_n(\dm) = \Ul_n$, except for no defense.
    With this in mind, we investigate the following defense mechanisms: 
    
    1. No Defense (``$\dno$''): the perturbation $e_i=0$ for $i=1,\ldots, n$. Note that $\dno$ has zero utility loss, differently from other defenses that incur a positive utility loss. 
    
    2. IID Noising (``$\drn$''): $e_i$'s are IID Gaussian with mean zero and variance $\sigma^2_n = \Ul_n > 0$. 
    
    3. Constant Noising (``$\dun$''): $e_i=\tau_n, i=1,\ldots, n$, where $\tau_n=\pm \sqrt{\Ul_n}$ is a constant for each $n$.

    These three defense mechanisms serve as natural baselines and are named here accordingly.
    To our knowledge, the literature has not yet provided formal definitions of such defense mechanisms.
    Guided by the developed theory, we also propose novel defenses that are rate optimal under some conditions. Throughout this section, we focus on regression tasks unless stated otherwise. 
    % Our findings are summarized in \Autoref{tab:def}. 
    % The proofs of all theorems and propositions are included in Appendix. 

    There are two unique challenges in the theoretical development of model privacy. The first challenge pertains to non-IID observations. Classical learning theory typically assumes that observations are IID or at least independent; however, the perturbations injected into model responses can have an arbitrary dependence structure. Therefore, traditional probabilistic tools that rely on independence are not applicable. This necessitates the development of new analytical tools. 
    The second challenge involves the allowance for diminishing noise levels. Classical learning theory usually assumes a fixed noise level, but a vanishing noise is of particular interest under model privacy.
    Specifically, a vanishing noise means that the service quality remains almost unaffected for benign users, which is especially appealing to the model owner. Our findings indicate that it is possible for the model owner to defend against model stealing attacks while maintaining high service quality.

    \textbf{Setup.} Unless specified otherwise, throughout the analysis in \Autoref{sec:model_func, sec:atk}, we assume the defender, attacker, and benign users all use the squared error loss. Specifically, $\loss_A(f,g) = \loss_D(f,g) = \norm{f-g}_2^2$ for any two functions $f$ and $g$, where the norm is with respect to query distribution $\mu$, which will be introduced soon. Additionally, for any two scalars $x$ and $y$, we have $\loss_U(x,y) =  (x-y)^2$. As explained in \Autoref{subsec:quantify}, the defenses under comparison should be within the same level of utility loss. Therefore, the defense mechanism set under consideration is $\dmc(\Ul_n) = \{\dm: \ul_n(\dm) \leq \Ul_n\}$, where $\{\Ul_n,n=1,\dots\}$ is a sequence of non-negative utility loss levels such that $\Ul_n \lesssim 1$. 
    % In contrast to querying the defender, the attacker may obtain ``nature-collected'' data. We assume that those natural data contain IID noise with mean zero and a fixed variance $\sigma^2>0$, and its distribution follows $\mu$ as well.
        The following assumptions are frequently used throughout the analysis in \Autoref{sec:model_func, sec:atk}. 
        \begin{assumption}[Same Query Behavior]\label{asmp:same_query}
            Both benign users and the attacker will send IID batch queries sampled from an absolutely continuous distribution $\mu$ on $\Real^d$. We denote IID batch query strategy as $\qiid$. 
            % The weights of utility loss (\Autoref{def:util}) are chosen as $w_i=1/n$ for $i=1,\dots,n$.
        \end{assumption}
        
        \begin{assumption}[Bounded Input Space] \label{asmp:bounded_input}
            The input space $\mathcal{X} = [0,1]^d$.
        \end{assumption}

        \begin{assumption}[Positive Density]\label{samp:density}
            The density of $\mu$ satisfies $\mu(x) \geq c$ for all $x \in [0,1]^d$ with some constant $c>0$.
        \end{assumption}

    % \begin{table}
    %     \centering
    %     \resizebox{\textwidth}{!}{
    %     \begin{tabular}{cccccc}
    %         \toprule
    %         \multicolumn{1}{c}{\multirow{2}{*}{Defense Mechanism}} &\multicolumn{1}{c}{\multirow{2}{*}{}}
    %         & \multicolumn{3}{c}{Attacker's Learning Algorithm} \\ 
    %         \cmidrule(l){3-6}
    %          &  & $k$-Nearest Neighbor & Polynomial Rgression & Kernel Ridge Regression & Neural Networks \\ \midrule 
    %        \multirow{2}{*}{No Defense}  & Potent (\Autoref{def:defstr}) & \xmark &\xmark&\xmark & \xmark \\ 
    %          & Rate optimal (\Autoref{def:priv}) &NA&NA&NA & NA\\ \hline
    %         \multirow{2}{*}{Constant Noising}  & Potent & \cmark &\cmark&\cmark&  \cmark\\ 
    %          & Rate optimal &\cmark&\xmark&\xmark & \xmark\\ \hline
    %          \multirow{2}{*}{Order Disguise (proposed)}  & Potent &&\cmark&&  \\ 
    %          & Rate optimal &  & \textbf{?} & & \\ \hline
    %          \multirow{2}{*}{Kernel Confusion (proposed)}  & Potent &&&\cmark&\cmark  \\ 
    %          & Rate optimal &  &  &\cmark &\cmark \\ 
    %          \bottomrule
    %     \end{tabular}}
    %     \caption{Goodness of defense mechanisms under different attack scenarios. Order Disguise denotes our proposed defense mechanism against attackers using polynomial regression, and Kernel Confusion denotes our proposed defense against attackers with kernel ridge regression or wide neural networks. NA means not applicable, and a question mark indicates that the result remains unclear.}
    %     \label{tab:def}
    % \end{table}

    \def\knn{\textrm{knn}}
    \def\lip{\textrm{Lip}}
    \subsection{\texorpdfstring{$k$}{k}-Nearest Neighbor}
        The $k$-nearest neighbor algorithm \citep[see, e.g., ][]{gyorfi2002distribution}, denoted by $\la_k$, is a classical learning method that predicts the output at an input $x$ using the average value of the observed responses near $x$. Specifically, the observations are sorted in ascending order of $\norm{\X_i- x}_2$ as $(\X_{(1,n)}(x), \Y_{(1,n)}(x)), \ldots, (\X_{(n,n)}(x), \Y_{(n,n)}(x))$, and
        the fitted regression function is given by
        \begin{align}\label{eq:knn}
            \hat{f}_n(x) = k^{-1} \sum_{j=1}^k \Y_{(j,n)}(x),
        \end{align}
        where $k$ is a hyper-parameter decided by the attacker and may depend on $n$.
        In our analysis, the collection of attacker's possible learning algorithms is $\lac_{\knn} = \{\la_k: k>0 \}$. For $C>0$ and $0< \alpha \leq 1$, a function $f$ is called $(C,\alpha)$-H\"{o}lder continuous if 
        $\abs{f(\x) - f(\x')} \leq C \norm{\x - \x'}_2^\alpha$ holds for any $\x,  \x' \in [0,1]^d$.
        Let $\hld(C,\alpha)$ be the set of all $(C,\alpha)$-H\"{o}lder continuous functions. We assume that $\fd \in \hld(C,\alpha).$ 
        The requirement of $\alpha \leq 1$ is common in the $k$-NN literature, and it is known that $k$-NN achieves sub-optimal learning rates for smooth function classes~\citep{devroye2013probabilistic, Gadat2016class, zhao2021minimax}.
        \Autoref{thm:knn} below presents the privacy level of different defense mechanisms.

       \begin{theorem}\label{thm:knn}
           Suppose that $\Ul_n \dgeq n^{-2\alpha/d}$.
           % $\lim_{n\to\infty} n^{2\alpha/d}\Ul_n =\infty $.
           Under \Autoref{asmp:same_query, asmp:bounded_input, samp:density}, the following hold:
           
                (i) For $\dno$, we have $\pl_n(\dno,\hld(C,\alpha)\mid\qiid, \lac_{\knn} ) \lesssim n^{-2\alpha/d}$.
                
                (ii) For $\drn$, we have $\pl_n(\drn,\hld(C,\alpha)\mid\qiid, \lac_{\knn}) \asymp (\Ul_n/n)^{2\alpha/(d+2\alpha)}$.
               
                (iii) For $\dun$, we have $\pl_n(\dun,\hld(C,\alpha)\mid\qiid, \lac_{\knn}) \asymp \Ul_n$.
      
                (iv) For any defense mechanism $\dm \in \dmc(\Ul_n)$,
                we have $\pl_n(\dm,\hld(C,\alpha)\mid\qiid, \lac_{\knn}) \lesssim \Ul_n$.
        \end{theorem}
        Based on \Autoref{thm:knn}, we can assess each defense's effectiveness and answer questions posed in \Autoref{sec:motivation}. Three key observations are summarized below, where we may omit the attack strategy and function class in the notations for ease when there is no ambiguity.
        \textbf{1: Defense is necessary.} 
        Without defense, we have $\pl_n(\dno) \lesssim n^{-2\alpha/d}$, meaning that the needed sample size $n$ to build an $\hat{f}_n$ that is $\epsilon$-close to $\fd$ is at most at the order of $\epsilon^{-d/2\alpha}$. Instead of querying the defender, suppose that the attacker can collect IID data with a fixed noise variance $\sigma^2$ from nature. Then, the needed sample size $n$ to construct an $\epsilon$-close function $\hat{f}_n$ is at the order of $\epsilon^{-(d+2\alpha)/2\alpha}$. 
        As a result, the attacker has a strong motivation to steal from a defender without defense, because querying the defender significantly reduces the attacker's needed sample size compared to collecting data from nature.
        \textbf{2: Existence of more effective defense than IID Noising.} Constant Noising $\dun$ is a better defense mechanism compared to IID Noising. Particularly, when $\dun$ is deployed with $\Ul_n = \sigma^2$, the attacker cannot even rebuild a function that converges to $\fd$.
        \textbf{3: Identification of rate optimal defense.} $\dun$ is rate optimal because the fourth result of \Autoref{thm:knn} shows that $\pl_n(\dm) \lesssim \Ul_n$ for any defense mechanism $\dm \in \dmc(\Ul_n)$, and $\dun$ achieves this rate.

    \def\preg{\textrm{poly}}
    \subsection{Polynomial Regression}\label{subsec:lr}
        This subsection studies the scenario where the attacker's learning algorithm fits a polynomial function with model selection. 
        For simplicity, we assume that $\X$ is univariate. 
        The attacker considers the following nested polynomial function class:
        \begin{align*}
            \mathcal{G} = \biggl\{f_q(\cdot;\bbeta_q): x \to \phi_q(x)^\T \bbeta_q, \phi_q(x)=(1, x^{1},  \ldots, x^{q})^\T,  \bbeta_q \in \Real^{q+1}, q=0,1,\dots, q_n \biggr\},
        \end{align*}
        where $q_n<n$ is the highest order of polynomials considered for each sample size $n$.
        To perform model selection, the attacker uses the generalized information criterion $\AIC$ \citep[see, e.g., ][]{shao1997asymptotic, ding2018model}.
        Specifically, for a given hyper-parameter $\lambda_n$ for model complexity regularization, the attacker fits the following function:
        \begin{align*}
            \hat{f}_n = \argmin_{f_q(\cdot;\bbeta_q) \in \mathcal{G}}   \biggl[ \frac{1}{n}\sum_{i=1}^n \{\Y_i-f_q(X_i;\bbeta_q)\}^2 + \frac{ \lambda_n \hat{\sigma}_{n,q}^2 q}{n} \biggr],
        \end{align*}
        where $\hat{\sigma}_{n,q}^2$ is an estimator of the variance of the random error in the regression model, assumed to be upper bounded by a constant $B>0$. The collection of possible learning algorithms is $\lac_{\preg} = \{\la_{\lambda_n}: \lambda_n \geq 2, \lambda_n / n \to 0\}$.
         We assume that the defender's model $\fd$ lies in a family $\fc_{\preg}=\{f: x \to \phi_p(x)^\T\bbeta_p, \bbeta_p \in \Real^{p+1}, \beta_{p+1} \neq 0, \norm{f}_2\leq C \}$ with a fixed order $p$ and constant $C>0$. 

        % \begin{assumption}\label{asmp:pr}
        % There exist some  positive constants $c_1,c_2,c_3,$ and $B$ such that
        %     (1) $\lim_{n\to\infty} \sigma_n^2 e^{nc} = \infty$ for any $c>0$, and there exists a $\gamma<1/(4p+2)$ such that $\P(\abs{\hat{\sigma}^2_{n,q}-\sigma^2_n} \geq \gamma \sigma^2_n) \leq e^{-nc_3}$ for all $0<q<n$. (2) $\norm{f}_2^2 \leq B $ for every $f \in \fc$. 
        %         (3) $\P(\norm{\Sigma_q\hat{\Sigma}_q^{-1}} \geq c_1+c_2\ln(1/\epsilon)) \leq \epsilon$ for all $0<q<n$ when $n$ is sufficiently large, where $\Sigma_q=\E[\phi_q(X) \{\phi_q(X)\}^T]$ is the covariance matrix and $\hat{\Sigma}_q$ is its empirical estimate.
        % \end{assumption}

        % \begin{remark}[Discussion on technical assumptions]
        %     In \Autoref{asmp:pr}, the first condition says that utility loss is not decreasing too fast, and the variance estimator is reasonably close to the truth; the second condition bounds $\fc$, the function class under consideration; and the last condition is typically met when the input $\X$ is bounded or follows a sub-Gaussian distribution. 
        % \end{remark}
        
        % \begin{theorem}[Polynomial regression]\label{thm:linear_reg}
        %     In the aforementioned scenario, the following results hold under \Autoref{asmp:pr}.
                
        %         (i) For $\dno$, we have $b_n(\dno) \lesssim e^{-nc}$ for some positive constant $c$.
                
        %         (ii) For $\drn$, we have $b_n(\drn) \asymp p\sigma_n^2/n$ under mild assumptions, as detailed in the appendix.
                
        %         (iii) For $\dun$, we have $b_n(\dun) \asymp \sigma_n^2$. 
        % \end{theorem}

        In this scenario, we propose a defense mechanism named Order Disguise ($\dour$). It can mislead the attacker to overfit the underlying model, significantly amplifying the privacy level compared to utility loss. $\dour$ involves constructing two perturbation directions $\e_1$ and $\e_2$ determined by the queries. 
        Here, $e_2$ is generated by a $k$-th order polynomial function with $k > p$. When the magnitude (in terms of $\ell_2$-norm) of $e_2$ is sufficiently large, our analysis shows that a $k$-th order polynomial minimizes the attacker's penalized empirical loss. Consequently, rather than recovering the true $p$-th order model $\fd$, the attacker overfits to this misleading higher-order structure introduced by $e_2$. Meanwhile, the construction of $\e_1$ ensures that this overfitted model performs poorly.
        % While $\e_2$ traps the attacker into overfitting, $\e_1$ ensures that the fitted model performs poorly. 
        The final perturbation $\e$ is a combination of $\e_1$ and $\e_2$, preserving the advantages of both.
        We formalize this result in \Autoref{thm:pos} and summarize the detailed steps of the proposed defense in \Autoref{alg:poly}. 

        \begin{algorithm}[tb]
        \caption{Defense Mechanism ``Order Disguise'' ($\dour$) Against Polynomial Regression}\label{alg:poly}
        \begin{algorithmic}[1]
            \Require Defender's model $\fd$, queries $X_i, i=1, \ldots ,n$, budget of utility loss $\Ul_n$, polynomial order to mislead attacker (target order) $k \in [p, q_n)$
            \State $\Phi_k = (\phi_k(X_1), \ldots, \phi_k(X_n))^\T \in \Real^{n\times(k+1)}$, $\u=(0,\ldots,0,1)^\T \in \Real^{k+1}$
            \State $\e_1 = \Phi_k \u$, $\e_2 = \Phi_k (\Phi_k^\T \Phi_k)^{-1} \u$ 
            \State $\e = \e_1/\norm{\e_1}_2+\e_2/\norm{\e_2}_2$
            % \State $\tilde{\e_i} = \e_i/\norm{\e_i}_2, i=1,2$
            % \rv{\State $\e = \tilde{\e_1} + \sign(\tilde{\e_1}^\T \tilde{\e_2}) \tilde{\e_2}$} \Comment{$\sign(x)=1$ if $x>0$, otherwise $-1$}
            \State $\e = \sqrt{n \Ul_n} \e / \norm{\e}_2$ \Comment{Normalize to the desired utility loss level}
            \Ensure $\Y_i = \fd(X_i) + e_i, i=1, \ldots ,n$
        \end{algorithmic}
        \end{algorithm}    
        
        \begin{theorem}\label{thm:pos}
        Suppose that (i) $n \cdot \P(\abs{X} \geq [\E (X^{2q_n}) / q_n^\gamma ]^{1/(2q_n)}) \to 0$ as $n\to\infty$ for a constant $\gamma > 0$, and (ii) $\Ul_n \geq 2 B q_n \lambda_n  /n$. 
        Under \Autoref{asmp:same_query}, when the defender adopts Order Disguise ($\dour$), there exists a sequence $\{k_n, n=1,2,\dots \}$ with $k_n \in [p, q_n)$ and $\lim_{n\to\infty}k_n=\infty$ such that the attacker will fit a function $\hat{f}_n$ with the highest order as $k_n$, and $\pl_n(\dour,\fc_{\preg}\mid\qiid,\lac_{\preg}) \gtrsim k_n^\gamma \Ul_n.$ 
        As a specific case, when $X$ follows a Gaussian distribution and $q_n\gtrsim \ln n$, we can take $\gamma=2$ and $k_n=4\ln n$.
        Another example is when $X$ follows a Beta distribution $Beta(\alpha,\beta)$ with $\beta>1$ and $q_n\gtrsim n^{1/\beta}$, we can choose any $\gamma > 0$ and $k_n=n^{1/\delta}$ for any $\delta \in (1, \beta)$.  
        \end{theorem}
        % \begin{remark}[High probability]
        % \Autoref{thm:pos} imposes a mild requirement that the tail of $X$ should decay fast enough. It is not hard to verify many common distributions such as Gaussian, Exponential, and Student $t$-distributions satisfy that condition with any fixed $\gamma > 0$. 

        \Autoref{thm:pos} implies that by using the proposed Order Disguise defense mechanism, the defender can achieve a high privacy level while maintaining a small utility loss. The gain-to-loss ratio $\pl_n/\Ul_n$ can even approach infinity. 
        % The key reason for the high privacy is that $\dour$ introduces \textbf{query-specific perturbations} targeting attackers using polynomial regression. 
        In contrast, without defense, $\pl_n(\dno)$ is zero for sufficiently large $n$. This implies that $\pl_n(\dun)\asymp \Ul_n$ under the Constant Noising defense because the attacker will learn the underlying function plus the injected constant as a whole. Also, the classical linear regression theory indicates that the privacy level under IID Noising is typically at the order of $U_n/n$. 
        While IID Noising and Constant Noising inject query-independent perturbations, Order Disguise utilizes the knowledge that the attackers use polynomial regression and carefully designs \textbf{query-specific perturbations} targeting such attackers, thus achieving a higher privacy level.
        % Therefore, with a similar analysis to the $k$-NN scenario, we know that (1) defense is necessary for the defender; (2) constant noising $\dun$ is a \eco defense but has a lower privacy level than $\dour$, hence not rate optimal.

        % \begin{remark}[Attacker that is Easier to Defend Against]
        %     \Autoref{thm:pos} together with \Autoref{thm:knn} suggest that an attacker using polynomial regression is easier to defend against than that using $k$-NN. \Autoref{thm:knn}(iv) implies that it is impossible to achieve a significantly higher privacy level than utility loss when the attacker uses $k$-NN. In contrast, with polynomial regression, the attacker can be misled to severely overfit the model, resulting in a large privacy level.
        %     We conjecture that an attacker who is more manipulable, meaning their learned model is more sensitive to the training data, is easier to defend against. We further elaborate on this idea in \Autoref{subsec:manipulate}.
        % \end{remark}
        
        \begin{remark}[Discussion on Technical Conditions]
        The first condition in \Autoref{thm:pos} ensures that there exists a direction where the overfitted model will have a large prediction error. It holds when the tail of $X$ decays sufficiently fast and $\lim_{n\to\infty} q_n \to \infty$. For example, the tail of Beta distribution $Beta(\alpha,\beta)$ is dominated by the second parameter $\beta$, and a larger $\beta$ indicates a faster speed of decay. Thus, we can guarantee the first condition when $\beta > 1$ and $q_n\gtrsim n^{1/\beta}$.
        The second condition on $\Ul_n$ ensures that the perturbation is large enough to cause the attacker to overfit, which is reasonably mild since $\lim_{n\to\infty} q_n\lambda_n/n \to 0$ is common in practice.
        % \rv{Lastly, Theorem~\ref{thm:pos} still holds when the defender has a polynomial function with an arbitrary high order, instead of a given order $p$. 
        % Recall that the privacy level (\Autoref{def:priv}) takes supremum over the defender's function class, the privacy level of $\dour$ is therefore non-decreasing when $\fc_{\preg}$ is enlarged to contain all polynomial functions. With that said, under $\dour$, the attacker using polynomial regression will always be misled to overfit regardless of $\fd$'s order.}
        \end{remark}

        \begin{remark}[Discussion on Model Mis-specification]
        In this section, we implicitly assume that the attacker correctly specifies the defender's function class $\fc$. 
        In general, a mis-specified function class of the attacker's learning algorithm typically benefits the defender. For example, in our polynomial regression scenario, both $\fd$ and $T$ come from parametric model classes. 
        A mis-specified class of $T$, such as two-layer neural networks with a fixed layer width, implies that there exist some $\fd$ that cannot be exactly recovered by the attacker even without defense. 
        In such cases, the attacker’s estimation error persists due to mis-specification, thereby reducing the urgency of defending against such attackers.
        However, since in practice the defender does not know the attacker's learning algorithm, effective defense mechanisms are still essential to protect against potential stealing attacks.
        \end{remark}

\section{Defending against an Unknown Attack Learning Algorithm} \label{sec:atk}
    
In this section, we study the defense against an attacker whose learning algorithm is unknown to the defender. Specifically, we consider a regression setting where the attacker aims to steal a H\"{o}lder smooth function and a classification setting where the target is a binary classification function. All proofs are included in the supplementary material. In both settings, our results address the minimum guaranteed privacy level provided by three defense mechanisms.

% Our results address the minimum guaranteed privacy level provided by three defense mechanisms.

\subsection{Regression with a H\"{o}lder Function Class}
We consider a regression setting where the defender's function $\fd$ belongs to a H\"{o}lder class $\hld(C,\alpha)$ with $C>0$ and $\alpha \in (0, 1]$. The attacker's learning algorithm collection $\lac$ includes any measurable mappings from query-response pairs to a measurable function of $\X$. The privacy level of three defenses is analyzed in the following: IID Noising, Correlated Noising, and No defense.

\textbf{IID Noising Defense.}
    We first study the IID Noising defense $\drn$, where the defender adds IID Gaussian noise with a common variance $\sigma_n^2$
    % \todo{may extend to other distribution than Gaussian} 
    to the responses. Its privacy level against an arbitrary attacker is derived as follows.

     \begin{theorem} % [H\"{o}lder class, IID noise]
     \label{ex:holderIID}
        Under \Autoref{asmp:same_query, asmp:bounded_input, samp:density},
        we have
        \begin{align*}
            \pl_n(\drn,\hld(C,\alpha) \mid\qiid,\lac) \begin{cases} 
            \asymp  (\sigma_n^2/n)^{2\alpha/(2\alpha+d)}, & \sigma_n^2 \gtrsim n^{-2\alpha/d} \\ 
            \lesssim n^{-2\alpha/d} , & \sigma_n^2 \lesssim n^{-2\alpha/d}.
            \end{cases}
        \end{align*}
    \end{theorem}

    Notably, \Autoref{ex:holderIID} \textit{allows the noise variance $\sigma_n^2$ to vanish}. When the noise variance is fixed, the privacy level can be analyzed using classical minimax theory. However, the case where $\sigma^2_n \to 0$ as $n \to \infty$ has received little attention in the existing literature. Gaining an understanding of this case is crucial for model privacy, as a defense with vanishing noise can still be successful, meaning that the attacker cannot steal a good model while the benign users' service quality is well maintained. Technically, to handle the case with vanishing noise, we need to derive the exact convergence rates of prediction error brought by randomness in queries and perturbations, respectively. 

    \Autoref{ex:holderIID} assumes that the attacker adopts an IID batch query strategy. 
    % We can prove that IID batch query is indeed optimal for the attacker. 
    The following theorem implies that in this regression task, IID batch query is the best choice for the attacker when $\sigma_n^2 \gtrsim n^{-2\alpha/d}$.
    
        \begin{theorem}[Non-IID Queries]\label{thm:IIDquery}
         Under \Autoref{asmp:same_query, asmp:bounded_input, samp:density}, we have $\pl_n(\drn,\hld(C,\alpha)\mid \qs,\lac) \gtrsim (\sigma_n^2/n)^{2\alpha/(2\alpha+d)}$ for every query strategy $\qs$. 
        \end{theorem}    
        % For any $p>0$, we denote the $\epsilon$-covering entropy of the function class $\mathcal{F}$ under $L_p$ loss as $V_p(\epsilon)$ \citep[see, e.g.][]{yang1999information}. 
        % \begin{assumption}\label{asmp:atk_iid}
        %     1. $\fc$ is rich, that is, $\liminf_{\epsilon \to 0^+}S(\alpha\epsilon)/S(\epsilon) > 1$ for some $\alpha \in (0,1)$. 2. There exist positive constants $a$, $b$, and $c$ such that $S(\epsilon) \leq V(b\epsilon) \leq cS(\alpha\epsilon)$ when $\epsilon$ is sufficiently small. 
        %     3. The noise variance $\sigma_n^2 \lesssim 1$.
        % \end{assumption}

        % Let $R_n(\qiid)$ and $R_n(\qs)$ denote the stealing limit with respect to IID batch query strategy and an arbitrary query strategy $\qs$, respectively.
        % Let $\pl_n(\drn,\fc\mid \qs,\lac)$ denote the privacy level of $\drn$ under an arbitrary query strategy $\qs$.
        % \begin{align*}
        %     R_n(\fc,\dmc,\qiid,\lac) \asymp R_n(\fc,\dmc,\qsc,\lac).
        % \end{align*}
        % Compared to \Autoref{ex:holderIID}, \Autoref{thm:IIDquery} implies that in this regression task, an attacker using IID batch query achieves the best stealing error rate when $\sigma_n^2 \gtrsim n^{-2\alpha/d}$.
        % However, the choice of query strategy has to be carefully considered for classification tasks, and IID query strategy is typically suboptimal in this case. In particular, the behavior of a classifier near the decision boundary is most critical, thus the attacker may pay more attention to those areas. \todo{Need support, using logistic example or simply cite some work?}

    % \end{remark}

    \textbf{Correlated Noising Defense.}\label{subsec:corr_noise}
    Using an IID Noising defense mechanism is often insufficient to prevent model stealing attacks. To enhance protection, the defender can add noise with correlation.
    One such choice is adding stationary noise with variance $\sigma_n^2$. Specifically, we have $\Y_i = \fd(X_i) + e_i, i=1,\dots,n$ with $\E(e_i) = 0$, $\var(e_i)=\sigma^2$, and 
    the correlation between any two perturbations $e_i$, $e_j$ that $\abs{i-j}=k$ equals $r(k)$, $k \geq 1$ for some function $r$ on integers.
    For example, an IID Noising defense mechanism $\drn$ has $r(k)=0$; $\sum_{k=1}^\infty \abs{r(k)} < \infty$ represents short-range dependent noise; and a typical example of long-range dependent noise satisfies $r(k) \asymp k^{-\gamma} $ for some $ 0<\gamma<1$. \Autoref{ex:hld_longrange} analyzes the privacy level of adding long-range dependent noise.

   \begin{theorem}% [H\"{o}lder class, long-range dependent noise]
    \label{ex:hld_longrange}
        Let $\dm_{\gamma}$ represent a defense mechanism that adds long-range dependent noise with $r(k)\asymp k^{-\gamma}$ for some $\gamma \in (0,1)$ and $ \ul_n(\dm_{\gamma}) = \sigma_n^2$. 
        Under \Autoref{asmp:same_query, asmp:bounded_input, samp:density}, we have 
        \begin{align*}
            \pl_n(\dm_{\gamma},\hld(C,\alpha)\mid \qiid,\lac)  \begin{cases} 
            \asymp (\sigma_n^2/n)^{2\alpha/(2\alpha+d)}+\sigma_n^2n^{-\gamma}, & \sigma_n^2 \gtrsim n^{-2\alpha/d} \\ 
            \lesssim n^{-2\alpha/d} , & \sigma_n^2 \lesssim n^{-2\alpha/d}.
            \end{cases}
        \end{align*}
    \end{theorem}

    \Autoref{ex:hld_longrange} demonstrates that the correlation structure of noise can be tailored to impede the attacker's ability to learn the defender's model effectively. Compared to \Autoref{ex:holderIID}, adding long-range dependent noise instead of IID noise can significantly increase privacy level. Moreover, a stronger correlation (a smaller $\gamma$) in the noise leads to a higher privacy level.

    \begin{remark}[Influence of Query Strategy]
        Based on the results of \citet{hall1990nonparametric, wang1996function}, we can show that $\pl_n(\dm_{\gamma},\hld(C,\alpha)\mid \qs,\lac) \asymp n^{-2\alpha\gamma/(2\alpha+\gamma)}$ when $\qs$ is an equally-spaced fixed design, $\sigma_n^2 \asymp 1$, and $\X$ is one-dimensional.
        This rate is slower compared to using an IID batch query~\citep{efromovich1999overcome, yang2001nonparametric}. As a result, when the attacker uses a batch query strategy, the defender can add long-range dependent noise in the order of $X_i$'s values to hinder the attacker's ability to learn $\fd$. In contrast, if the attacker adopts a sequential query strategy that requests immediate responses, the defender can only add long-range noise in the order of queries, potentially allowing the attacker to learn $\fd$ at a faster rate of convergence. 
    \end{remark}

    \def\fixq{\textrm{Fix}}
    \textbf{No Defense.}
    To understand whether a defense is necessary, it is crucial to investigate the scenario where the defender does not apply any defense.
    Let $\dno$ denote the defense mechanism with no perturbation, under which $\Y_i=\fd(X_i)$. For technical convenience, we assume the benign user has a uniform query distribution. 
    Then, \Autoref{ex:atk_no} gives the privacy level of no defense against an attacker whose query strategies $\qsc_{\fixq}$ include all fixed designs. Here, a fixed design query strategy consists of pre-selected sets of data points $\X_1, \dots, \X_n$ for each sample size $n$. 
    \begin{theorem} % [H\"{o}lder class, no defense]
    \label{ex:atk_no}
       Under \Autoref{asmp:same_query, asmp:bounded_input, samp:density},
        suppose the attacker's query strategy collection $\qsc$ includes all fixed designs. Then, we have $\pl_n(\dno,\hld(C,\alpha)\mid \qsc_{\fixq},\lac) \asymp n^{-2\alpha/d}$. 
    \end{theorem}

    Compared to \Autoref{ex:holderIID}, where the defender adds IID noise, the privacy level is significantly smaller when there is no defense. This implies that attackers have a strong incentive to steal from a defender without defense. 
    % However, injecting long-range dependent noise with a small correlation parameter $\gamma$ into the responses can make it worthless for attackers to steal the model, as \Autoref{ex:hld_longrange} implies a much higher privacy level than IID Noising. 

\newcommand{\flin}{\mathcal{H}_{\textrm{lin}}}
\newcommand{\lalin}{\lac_{\textrm{lin}}}
\newcommand{\laerm}{\la_{\textrm{erm}}}
\newcommand{\hd}{h^*}

\subsection{Binary Classification}
As another illustrative example of model privacy, we study a binary classification task where the defender has a linear classifier $\hd: [0,1]^d \to \{0, 1\} \in \flin \defeq \{\ind_{x^\T\beta \geq 0}, \beta \in \Real^d \}$. That is, the defender will yield a hard label $1$ for a half space determined by $\x^\T \beta \geq 0$, and return label $0$ for the other half. In the following, we focus on defense mechanisms that also return hard labels instead of soft labels (e.g., the probability that a query takes label $1$). 

The attacker aims to reconstruct a binary classifier $\hat{h}_n: [0,1]^d \to \{0, 1\}$ that is close to $\hd$. 
Regarding the query strategy, we stick to \Autoref{asmp:same_query, asmp:bounded_input, samp:density} used in Section~\ref{sec:model_func}. Namely, the attacker sends IID batch queries generated from a distribution $\mu$ that has a bounded support on $[0, 1]^d$ and a positive density. The attacker's learning algorithm is unknown to the defender, and $\lalin$ denotes all measurable mappings from data to $\flin$.
Given hard labels, we adopt the zero-one loss function $\ell(h,g) = \E (\ind_{h \neq g})$ for defender, attacker, and benign users. Notably, the utility loss in this case is $\Ul_n = \E\bigl(n^{-1}\sum_{i=1}^n \ind_{\Y_i \neq \hd(\X_i)}\bigr) \in [0,1]$, measuring the fraction of responses that do not match $\hd(\X_i)$. 

Three baseline defense mechanisms analogous to the regression setting are considered. While No Defense ($\dno$) remains the same, IID Noising and Constant Noising defenses are redefined as follows. 

1. IID Noising ($\drn$): 
$\Y_i = \hd(\X_i)$ if $W_i = 1$; otherwise $\Y_i = 1 -\hd(\X_i)$, 
$i=1,\dots,n$, where $W_i$'s are IID Bernoulli random variables with $\P(W_i=1)=\Ul_n$. In other words, each label is randomly perturbed with probability $\Ul_n$. 

2. Constant Noising ($\dun$): $\Y_i = \ind_{\X_i^\T\beta + s_n \geq 0}, i=1,\dots,n$, where $s_n$ is a constant. The constant $s_n$ added to the score $x^\T\beta$ leads to a shift in the decision boundary of $\hd$. The utility loss $\ul_n(\dun)$ monotonously increases when $\abs{s_n}$ increases, which can therefore be controlled by the defender as the benign user's query distribution is often known or can be well-estimated in practice. 
% and the defender can properly choose $s_n$ to achieve a desired utility loss level.

    \begin{theorem}\label{thm:class}
           Under \Autoref{asmp:same_query, asmp:bounded_input, samp:density}, the following hold:
           
                (i) For $\dno$, we have $\pl_n(\dno,\flin\mid\qiid, \lalin) \asymp n^{-1}$.
                
                (ii) For $\drn$ with $\lim_{n\to\infty} \Ul_n < 1/2$, we have $\pl_n(\drn,\flin\mid\qiid, \lalin) \asymp n^{-1}$.
               
                (iii) For $\dun$ with $ n^{-1} \lesssim \Ul_n \leq 1$, we have $\pl_n(\dun,\flin\mid\qiid, \lalin) \asymp \Ul_n$.
        \end{theorem}

    Unlike the results in the regression setting, \Autoref{thm:class} implies that IID Noising defense is essentially equivalent to No Defense, given that the rate of perturbing the response is less than $1/2$. This is expected since classification is known to be easier than regression of learning the underlying conditional probability function. Meanwhile, a defense with $\Ul_n \geq 1/2$ implies that over half of the responses will be perturbed, leading to an unacceptable service quality degradation. Moreover, such a large utility loss budget allows the defender to return random labels that contains no information about $\hd$, so that the attacker cannot learn $\hd$ based on the perturbed responses solely.

    Nevertheless, Constant Noising still achieves a high privacy level. 
    Similarly to the regression setting, the attacker will be fooled by the constant perturbation injected into the responses and learn it, leading to a biased estimation and large misclassification error. While our theoretical analysis here focuses on the classic classification task where the responses are hard labels,
    Misleading Shift (\Autoref{alg:ms} in \Autoref{sec:exp}) proposed in our real-world experiments can be regarded as an extension of Constant Noising to return soft labels, which are the logit of scores after adding a constant perturbation. 
    Therefore, we expect Misleading Shift to have a similar performance as Constant Noising in \Autoref{thm:class}.

\section{Experimental Studies} \label{sec:exp}

To demonstrate the practical application of the proposed model privacy framework, we conduct experiments on both simulated and real-world datasets to corroborate our theoretical results. Specifically, our results indicate that a specialized defense mechanism is essential to thwart model stealing attacks. In addition to the widely studied regression tasks in the earlier sections, we also conduct experiments on classification tasks, which represent another vital application domain. Notably, our theoretical framework guides the implementation of defenses that significantly bolster model privacy across both task types. The full experiments details can be found in the supplementary materials, \rv{and the code is available at \url{https://github.com/KeyWgh/ModelPrivacy}.}
% \href{https://github.com/KeyWgh/ModelPrivacy}{here}}.

\def\sigmoid{\textrm{softmax}}

    \begin{figure}[tb]
        \centering
        \includegraphics[width=0.8\linewidth]{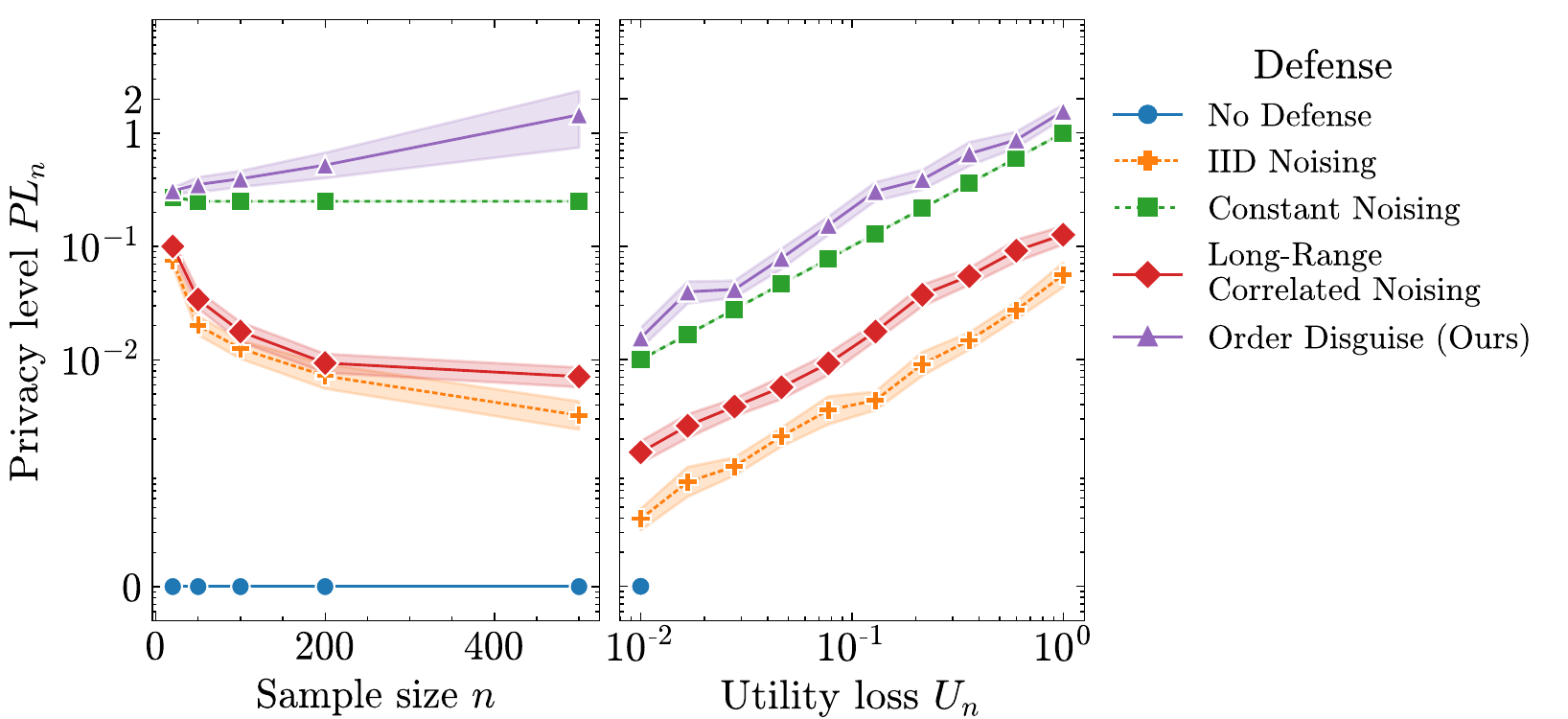}
        \caption[Goodness comparison of different defense mechanisms against an attacker using polynomial regression.]{Goodness comparison of different defense mechanisms against an attacker using polynomial regression. Left: Privacy level at different sample sizes with the utility loss level $\Ul_n=0.25$.  Right: Privacy level at different utility loss levels with the sample size $n=100$. The shaded area reflects one standard error.}
        \label{fig:compare_defense}
    \end{figure}

        \begin{figure}[tb]
            \centering
                \begin{minipage}{0.46\linewidth}
                    \centering                        \includegraphics[width=\linewidth]{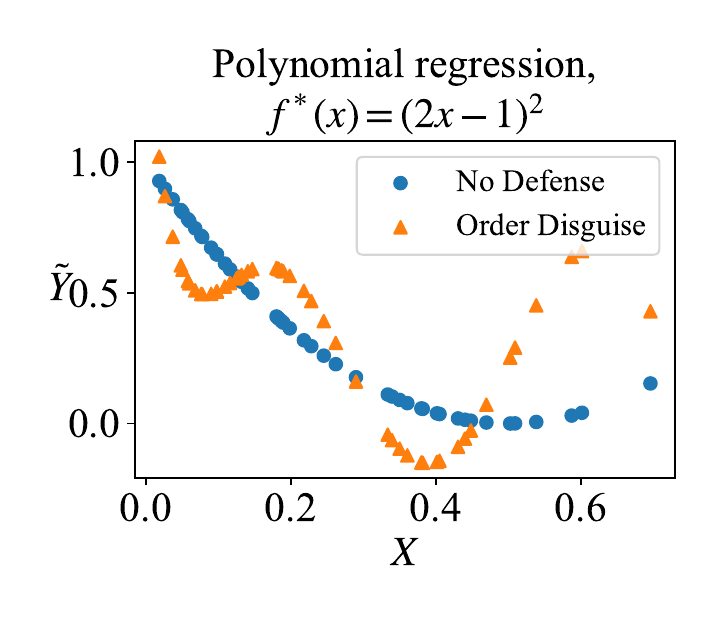}
                \end{minipage}
                \hfill
                \begin{minipage}{0.49\linewidth}
                    \centering
                        \includegraphics[width=\linewidth]{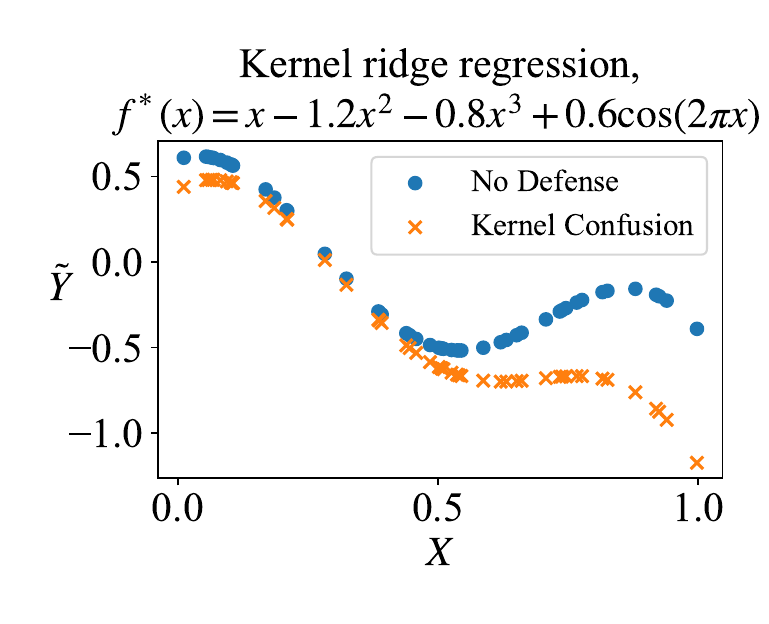}
                \end{minipage}
            \caption[The unprotected responses and perturbed responses by proposed defense mechanisms.]{
            Left: The unprotected responses and perturbed responses by Order Disguise (\Autoref{alg:poly}) against an attacker performing polynomial regression. Right: The unprotected and perturbed responses by Kernel Confusion (in the supplementary material Section~\ref{sec:supp_more_attacks}) against an attacker performing kernel ridge regression.
            % Examples showing the unprotected responses and perturbed responses under our proposed defense mechanism. Left: Using \Autoref{alg:poly} (`Order Disguise') to defend against an attacker performing polynomial regression. Right: Using a defense proposed in the supplementary material (`Kernel Confusion') to defend against an attacker performing kernel ridge regression.
            } 
            \label{fig:defense_example}
        \end{figure}    
        
\subsection{Simulated Experiments}
    
    \subsubsection{Polynomial Regression} 
    Suppose the defender owns a quadratic function $\fd(x) = (2x-1)^2$. The attacker performs polynomial regression with $q_n=n^{1/3}$ and selects the model using the Akaike information criterion (AIC) \citep{akaike1998information}. We also conduct experiments where the attacker adopts Bayesian information criterion (BIC) or Bridge Criterion (BC) \citep{ding2017bridging} as the selection criterion and found highly similar results compared to AIC. The squared error loss is used by both defender and attacker. 
    We examine five defense mechanisms: No Defense, IID Noising, Constant Noising (see \Autoref{sec:model_func}), Long-Range Correlated Noising (see \Autoref{subsec:corr_noise}), and our proposed defense Order Disguise (see Algorithm~\ref{alg:poly}).

    First, we study the privacy level of these defense methods under different sample sizes. For computational feasibility, the privacy level is evaluated on a function class containing a single function $\fd$ for all experiments in this section.
    Specifically, we vary the number of queries, $n$, with values set at $20, 50, 100, 200$, and $500$.
    For all defense mechanisms, the utility loss $\Ul_n$ is fixed at $0.25$. Equivalently, the signal-to-noise ratio, defined as $\E[\{\fd(\X)\}^2]/\E(e^2)$ with $e$ being the perturbation, is around one. 
    For each experiment, the attacker sends $n$ IID queries sampled from a Beta distribution $Beta(1,3)$ and receives responses from one of the five defense mechanisms. We then evaluate the privacy level of this defense mechanism using the test error of the rebuilt model on $1000$ IID test data drawn from $Beta(1,3)$. For every combination of sample size $n$ and defense mechanism, we run $100$ independent replicates.

    We also investigate the privacy level at different utility loss levels. Keeping all other experimental conditions constant, we conduct additional experiments with a sample size of $n=100$, varying the utility loss $\Ul_n$ from $0.01$ to $1$. 

    \textbf{Findings.}
    The results of both studies are summarized in Figure~\ref{fig:compare_defense}. 
    Clearly, the attacker can easily steal the defender's model when there is no defense, as the privacy level is zero with only 20 queries. Meanwhile, adding independent noise does not significantly improve the privacy level. The attacker can still efficiently learn a good model under IID Noising. Although Long-Range Correlated Noising performs better than IID Noising, its privacy level remains significantly lower than that of adding perturbations with attack-specific correlation patterns, such as Constant Noising and our proposed defense Order Disguise. 
    
    Moreover, Order Disguise is the only defense mechanism that achieves a higher privacy level than utility loss, aligning with \Autoref{thm:pos}. The left panel in \Autoref{fig:defense_example} demonstrates how Order Disguise injects dependent perturbations into the responses. The shape of the perturbed responses resembles a polynomial function with a higher order than the truth, supporting our theoretical finding that Order Disguise can mislead the attacker into severe overfitting. 

    In summary, the experimental results on polynomial regression align well with our developed theory in \Autoref{subsec:lr}, leading to three critical observations: (1) Defense is necessary; (2) Defenses using independent perturbations offer limited gains in the privacy level compared to no defense. (3) Defenses using dependent perturbations tailored to the attack scenario can be the most effective.

    \subsubsection{Penalized Regression with High-Dimensional Datasets}\label{subsub:high_dim}
        Suppose the defender owns a linear function $\fd(x) = \X^\T \bbeta$, where $\bbeta \in \Real^d$ is the model parameter. The attacker performs penalized regression with built-in variable selection methods, such as LASSO~\citep{tibshirani1996regression} and Elastic Net~\citep{zou2005regularization}. We consider the high-dimensional regression case where the number of predictors $d$ is larger or comparable to the number of queries $n$. We perform seven simulation examples including both sparse and dense structure of $\bbeta$. Similar results are observed across all simulation examples, thereby we describe simulation example 1 in detail as a demonstration. Full experimental details and results of other examples are included in Appendix. 

        Following the setting of \citet{zou2005regularization, nan2014variable}, simulation example 1 generates 50 queries with $d=40$. The model parameter $\bbeta$ is chosen as $\bbeta_i= 3\times \ind_{i\leq 15}$ for $i=1,\dots, 40$. The queries $\X_i = (\X_{i,1}, \dots, \X_{i, 40}), i=1,\dots,50$ are IID generated as follows: 
        $X_{i,j} = Z_{1} + \epsilon_{i,j}, Z_{1} \sim N(0, 1), j=1,\dots,5; 
            X_{i,j} = Z_{2} + \epsilon_{i,j}, Z_{2} \sim N(0, 1), j=6,\dots,10;
            X_{i,j} = Z_{3} + \epsilon_{i,j}, Z_{3} \sim N(0, 1), j=11,\dots,15;
            X_{i,j} \sim N(0, 1), j=16,\dots, 40. $
        % \begin{align*}
        %     X_{i,j} &= Z_{1} + \epsilon_{i,j}, Z_{1} \sim N(0, 1), j=1,\dots,5, \\
        %     X_{i,j} &= Z_{2} + \epsilon_{i,j}, Z_{2} \sim N(0, 1), j=6,\dots,10, \\
        %     X_{i,j} &= Z_{3} + \epsilon_{i,j}, Z_{3} \sim N(0, 1), j=11,\dots,15, \\
        %     X_{i,j} &\sim N(0, 1), j=16,\dots, 40. \\
        % \end{align*}
        Here, $Z_{1}, Z_{2}, Z_{3}$ are IID standard Gaussian, $X_{i,j}$ are IID for $j\geq 16$, and $\epsilon_{i,j}$ are IID $N(0, 0.01)$. 
        The penalty parameters for LASSO and Elastic Net are chosen with 5-fold cross-validation. 

\begin{figure}[tb]
        \centering
        \includegraphics[width=0.9\linewidth]{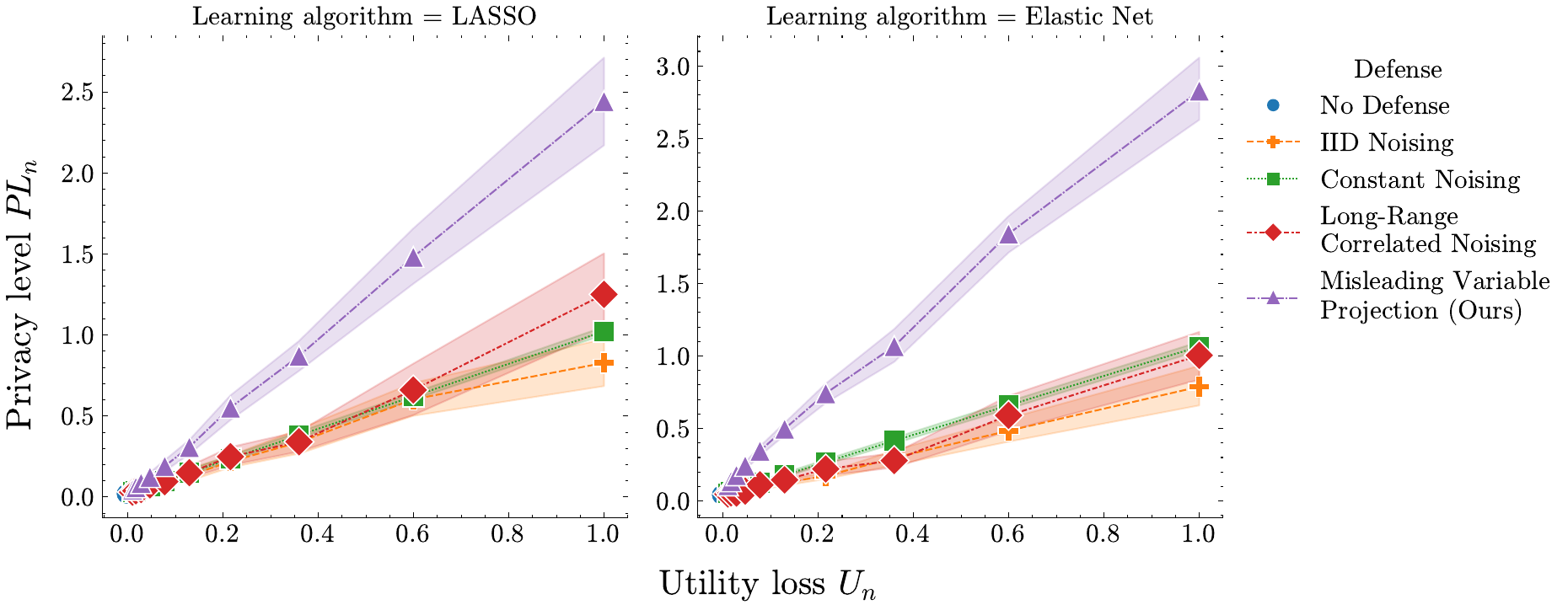}
        \caption[Goodness comparison of different defense mechanisms against an attacker performing penalized regression.]{Goodness comparison of different defense mechanisms against an attacker performing penalized regression. }
        \label{fig:compare_defense_lasso}
    \end{figure}

        Five defenses are used for comparison: No Defense, IID Noising, Constant Noising, Long-Range Correlated Noising, and our proposed method Misleading Variable Projection (abbreviated as MVP, see \Autoref{alg:lasso}). MVP is designed to mislead the attacker into fitting a model with non-significant variables. Specifically, the defender chooses a set of non-significant variables and adds perturbations to minimize the distance from $\hat{Y}$ to the space spanned by those variables.  This leads the attacker to incorrectly identify non-significant variables as significant, resulting in a mis-specified model and poor prediction performance. The privacy level is evaluated on $400$ test points, with results from 20 independent replicates presented in \Autoref{fig:compare_defense_lasso}. 

        \begin{algorithm}[tb]
        \caption{Defense Mechanism ``Misleading Variable Projection'' (MVP)}\label{alg:lasso}
        \begin{algorithmic}[1]
            \Require Defender's model $\fd$, queries $\X_i, i=1, \ldots ,n$, utility loss level $\Ul_n$, variable sampling ratio $\rho$
            \State Let $S=\{i: \bbeta_i \neq 0\}$ \Comment{The set of non-significant variables}
            \State Randomly sample and keep $\rho$ percent of $S$, and let $\abs{S}$ be its cardinality
            \State Let $\X_{i,S} \in \Real^{\abs{S}}$ be the random variable constrained on the index set $S$
            \State Let $\X_{S} = (\X_{1,S}, \dots, \X_{n,S}) \in \Real^{n\times \abs{S}}$, $Y^*=( \fd(\X_1), \dots, \fd(\X_n)) \in \Real^n$
            \State Let $\hat{Y} = \mathcal{P}_S Y^*$, where $\mathcal{P}_S = \X_{S} (\X_{S}^\T \X_{S})^{-1} \X_{S}^\T $ is the projection operator on the space spanned by $\X_{S}$
            \State Let $\u = \hat{Y} - Y^*$ and $c = \norm{\u}_2$ \Comment{Smallest distance from $Y^*$ to the space of $\X_{S}$}
            \State Let $\bv = Y^*/\norm{Y^*}_2$ \Comment{Normalize $Y^*$}
            \State $\e =  \sqrt{n} \Ul_n\u/c $ if $ \sqrt{n} \Ul_n \leq c$ else $\u + \sqrt{n\Ul_n^2 - c^2} \bv$ \Comment{Perturbation that push $\Y$ to the space of $\X_{S}$}
            \Ensure $\Y_i = \fd(\X_i) + e_i, i=1, \ldots ,n$
        \end{algorithmic}
        \end{algorithm}  
        
\begin{figure}
        \centering
        \includegraphics[width=0.9\linewidth]{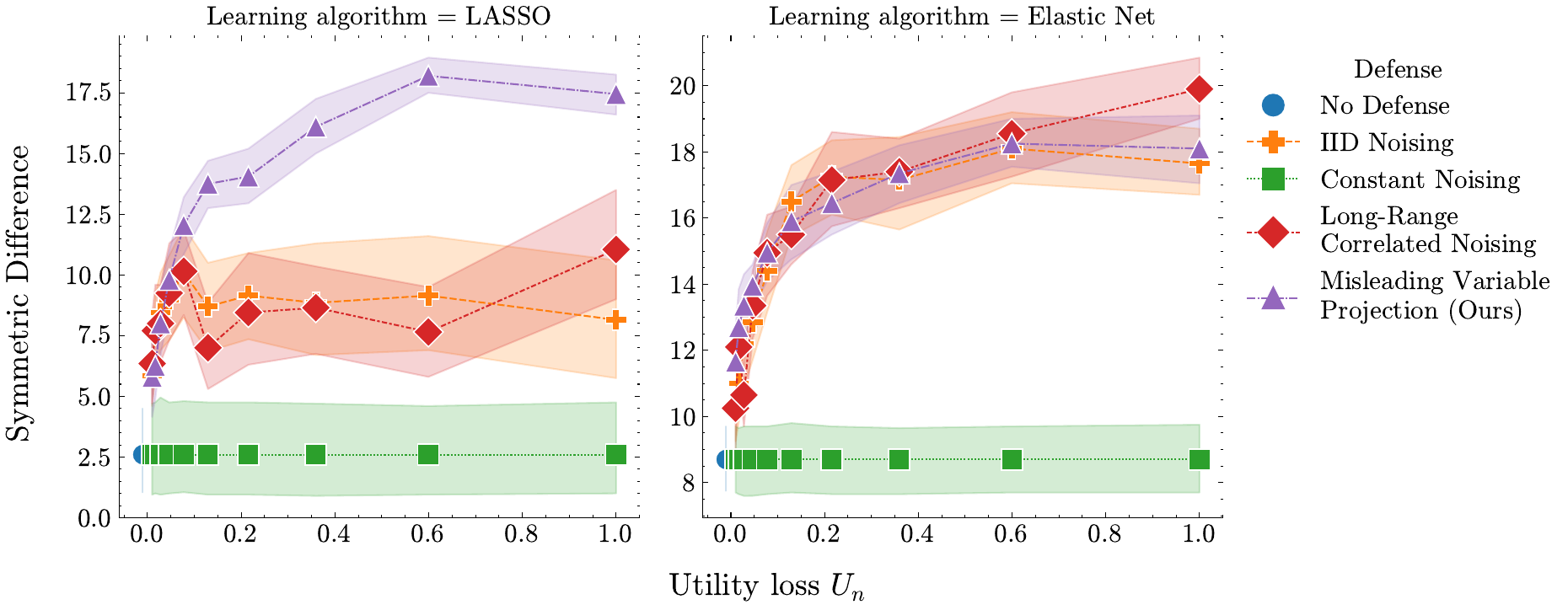}
        \caption[Variable selection reliability of different defense mechanisms against an attacker performing penalized regression.]{Variable selection reliability of different defense mechanisms against an attacker performing penalized regression. }
        \label{fig:compare_defense_lasso_diff}
    \end{figure}

    \textbf{Findings.}
    The results in \Autoref{fig:compare_defense_lasso} are consistent with previous experiments on the two regression tasks. Without defense, the attacker can rebuild a model with almost zero privacy level. Among all defenses, MVP provides the highest privacy level. Moreover, we empirically observe that MVP can deceive the attacker into selecting incorrect variables, as shown in \Autoref{fig:compare_defense_lasso_diff}. To evaluate variable selection accuracy, we calculate the symmetric difference between the true significant variable set $S^*\defeq \{i: \bbeta_i \neq 0\}$ and the set of variables selected by the attacker, denoted as $\hat{S}$. The symmetric difference is $\abs{S^*\backslash \hat{S}}+ \abs{\hat{S} \backslash S^*}$, where $\abs{S}$ is the cardinality of a set $S$. A smaller symmetric difference indicates a higher similarity between the two variable sets, thus a higher variable selection accuracy. From \Autoref{fig:compare_defense_lasso_diff}, we find that MVP leads to the highest symmetric difference.

% \subsection{A More Manipulable Attacker is Easier to Defend Against}
%     We support the findings in \Autoref{subsec:manipulate} by showing that an attacker with an enlarged function class is easier to defend against. In particular, with the same setup as the regression experiment in \Autoref{subsec:exp1}, but the attacker uses polynomial regression with the highest order $q_n$ takes values from $3,5,7,9$, and $11$. The privacy level significantly increases when $q_n$ increases, as presented in \Autoref{tab:lr_change_k}. 

%     \begin{table}
%             \caption{The empirical privacy level (standard error) when the attacker steals a quadratic function using polynomial regression with varied highest order $q_n$.}
%         \label{tab:lr_change_k}
%         \centering
%         \begin{tabular}{lccccc}
%         \toprule
%         $q_n$     &      3 &           5 &            7 &           9 &           11 \\ \midrule
%         Privacy level &   0.28 (0.01) &  0.3 (0.01) &  0.39 (0.03) &  0.59 (0.09) &  0.98 (0.19) \\
%         \bottomrule
%         \end{tabular}
%     \end{table}

\subsection{A real-world case study: hate speech detection}\label{subsec:nlp}
    We consider a model owner who has trained a large language model-based hate speech detection model. 
    Specifically, the defender's model $\fd$ is pretrained on the Toxic Comment Challenge Dataset~\citep{jigsaw-toxic-comment-classification-challenge}, which contains 159,566 sentences from Wikipedia's talk page edits, each labeled as either `normal' or `toxic'. For an input sentence $\X$, $\fd$ will output a predicted probability $\fd(\X) \in [0, 1]$ indicating the likelihood that the input is toxic. 
    
    An attacker, through querying the defender, aims to rebuild a detection function that performs comparably to $\fd$ at a small cost. In our experiments, the attacker's model $\hat{f}_n$ comprises a sentence BERT model~\citep{reimers2019sentence} that transforms the input sentence to an embedding vector, followed by a fully connected two-layer ReLU neural network that predicts the probability. The attacker's queries are sampled from the Toxic Comment Challenge Dataset or the Hate Speech Offensive Dataset~\citep{hateoffensive}, mimicking situations where the attacker either has knowledge of the true data distribution or can only use a surrogate dataset, respectively. The number of queries $n$ takes values of $1000, 2000$, and $5000$.

    % We adopt the same five defenses used in previous simulated experiments on classification tasks.
        Five defenses are studied in the experiments, as detailed below. It is worth noting that our proposed defense, Misleading Shift, is the only defense that adds dependent perturbations conditional on queries.

        1. No Defense, $\Y=\fd(X)$.

        2. Random Shuffle. We propose it as a counterpart of IID Noising for classification tasks. Since the response $\Y$ is a predicted probability vector, a natural way to perturb the response is to randomly shuffle all its coordinates. In particular, let $\Y=\fd(\X)$ with probability $1-\xi$; with probability $\xi$, $\Y$ takes one of the permutation of $\fd(\X)$'s coordinates with equal probability.

        3. Deceptive Perturbation proposed by \citet{lee2019defending}. It perturbs confident responses that are close to zero or one towards the opposite side (without changing the most likely class), making them less confident.

        4. Adaptive Misinformation proposed by \citet{kariyappa2020defending}. It first trains a confounding model that has low accuracy on the original task. Then, it replaces non-confident responses that are close to $0.5$ with the confounding model's outputs.

        5.  Misleading Shift. Inspired by the developed theory, we propose a method named Misleading Shift (see \Autoref{alg:ms}). The idea is to perturb all responses towards the direction of the dominating class among the queries, thereby misleading the attacker to overfit this dominating class and enhancing model privacy.

        \begin{algorithm}[tb]
        \caption{Defense Mechanism ``Misleading Shift'' for Classification Tasks}\label{alg:ms}
        \begin{algorithmic}[1]
            \Require Defender's model $\fd$, queries $\X_i, i=1, \ldots ,n$, scale parameter $\delta$, number of classes $K$
            \State Let $k \defeq \argmax_{1\leq j \leq K} \bigl\vert\{i: \fd(\X_i)_j = \max_{1\leq l\leq K} \fd(\X_i)_l \}\bigr\vert$ \Comment{\parbox[t]{.35\linewidth}{$\vert\cdot\vert$ is the cardinality of a set and $\fd(\X_i)_l$ is the $l$-th coordinate of $\fd(\X_i) \in [0,1]^{K}$}}
            \State  Let $c_k = (\ind_{k=1}, \dots, \ind_{k=K}) \in \{0,1\}^K$ \Comment{$\ind_{(\cdot)}$ is the indicator function}
            \Ensure $\Y_i = \sigmoid(\log(\fd(\X_i))+\delta c_k), i=1, \ldots ,n$
        \end{algorithmic}
        \end{algorithm}

\begin{figure}[tb]
    \centering
    \includegraphics[width=0.9\linewidth]{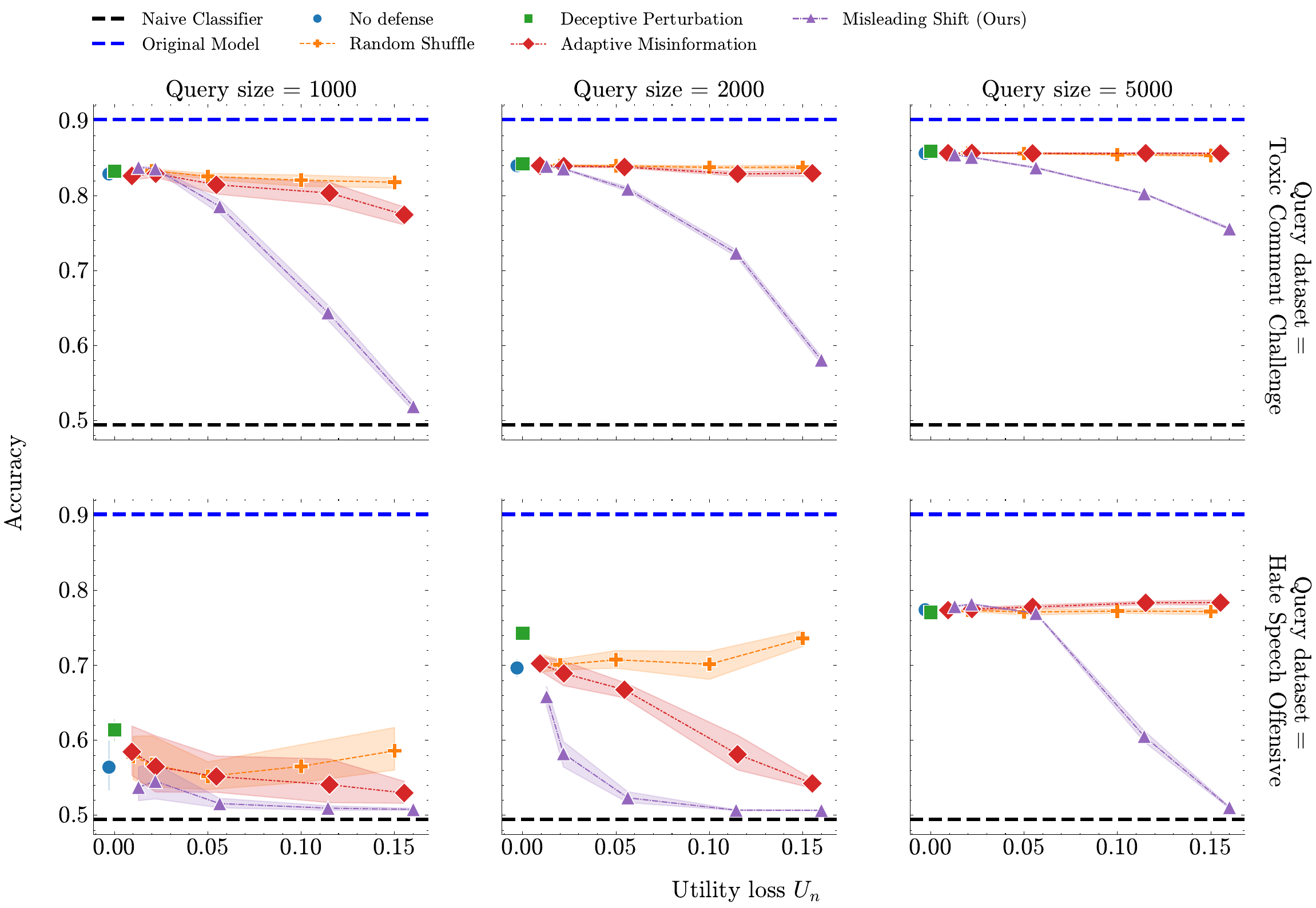}
    \caption[Average accuracy of the attacker's rebuilt model for the hate speech detection task.]{Average accuracy of the attacker's rebuilt model for the hate speech detection task. }
    \label{fig:nlp_acc}
    \vspace{-0.1em}
\end{figure}

\begin{figure}[tb]
    \centering
    \includegraphics[width=0.9\linewidth]{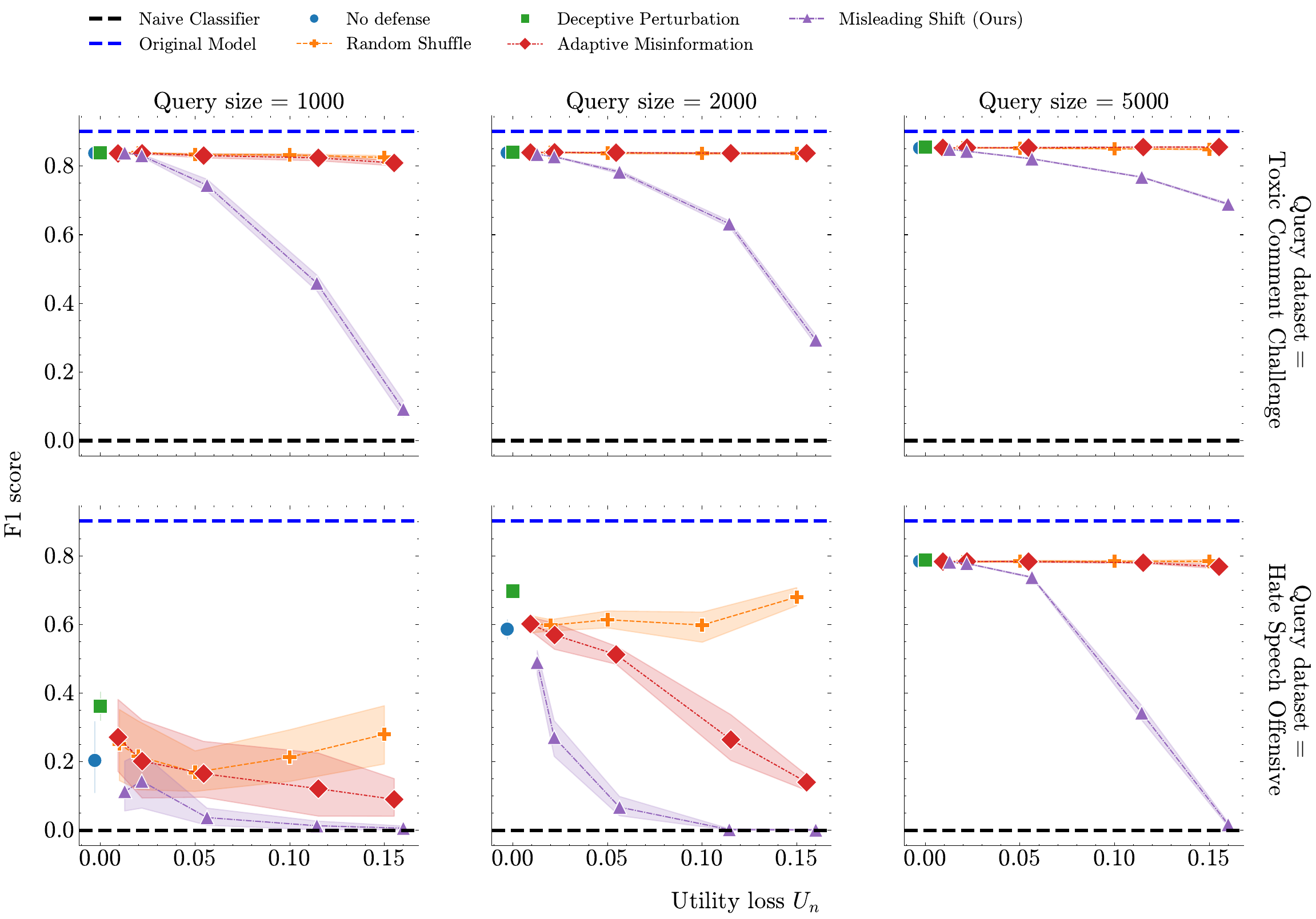}
    \caption[Average F1 score of the attacker's rebuilt model for the hate speech detection task.]{Average F1 score of the attacker's rebuilt model for the hate speech detection task.}
    \label{fig:nlp_f1}
\end{figure}

    As for the evaluation criterion, recall that the attacker aims to recover a model with high prediction performance in this real-world application. Therefore, we report the average accuracy and F1 score of the rebuilt model against the utility loss level of the defense over $10$ independent replicates. The accuracy and F1 score are obtained on a test dataset randomly sampled from the Toxic Comment Challenge Dataset. Moreover, we include two baselines for comparison: the accuracy and F1 scores of the defender's model $\fd$ (`Original Model') and a naive model that predicts zero or one with equal probability (`Naive Classifier'). The results are shown in \Autoref{fig:nlp_acc, fig:nlp_f1}.  

    \textbf{Findings.} From the experiment results, it is clear that defense is necessary to prevent stealing attacks. Suppose the cost of querying one sentence from the defender is one cent, and the cost of obtaining a human-annotated sentence is also one cent. Then, the defender's training cost of $\fd$ is over \$1000. However, the attacker can steal a model achieving 90\% of $\fd$'s performance for as little as \$10 without defense. Even with a surrogate dataset, the reconstruction cost is around \$50, which is just 5\% of the defender's expense.

    Also, the results show that adding independent perturbations conditional on queries does not significantly hinder the attacker. When the defender deploys Random Shuffle or Deceptive Perturbation, the attacker can still efficiently rebuild a good model as if there were no defense. Adaptive Misinformation also shows limited effectiveness, especially when queries come from the same distribution as the training data or when the number of queries is substantial.

    In a stark contrast, our proposed defense, Misleading Shift, achieves high privacy gains across all utility loss levels. Misleading Shift adds a constant, attack-specific perturbation to all responses. This attack-specific dependent structure of perturbations exaggerates the trend of the majority class in the queries, thereby consistently misleading the attacker to overfit the majority class. This insight is corroborated by \Autoref{fig:nlp_f1}, where the rebuilt model's F1 score is close to zero under misleading shift, indicating that the model predicts most inputs as the same class.

% \section{Discussions} \label{sec:dis}
    % \input{discussion.tex}

    % \todo{Add Conclusion section?}
\section{Conclusions} \label{sec:con}
    This paper establishes a framework called model privacy to understand and mitigate the vulnerabilities of machine learning models to model stealing attacks. By identifying key components of model stealing attacks, we quantify the goodness of stealing attacks and defense through statistical risk and formulate the objectives of both attackers and defenders.    
Within this framework, we have derived both theoretical and empirical results that yield valuable insights into effective defense strategies. Our findings highlight three key points. 
Firstly, defense is generally necessary for protecting models from being easily stolen. Without any defense, attackers can successfully reconstruct models across diverse architectures and attack scenarios. For instance, merely 1000 queries are sufficient to steal a large language model-based hate speech detector with 85\% accuracy, while the original model was trained on over 100,000 data points (see Subsection~\ref{subsec:nlp}). 
Secondly, defenses based on adding IID perturbations are typically ineffective. For such defenses, the level of service degradation for benign users often outweighs any improvement in the model's privacy level. 
Thirdly, the strategic disruption of the independence in perturbations is vital for enhancing model privacy. To demonstrate this, we have proposed novel theory-inspired defenses for both regression and classification tasks that significantly impair an attacker's ability to steal models. 
In conclusion, the model privacy framework offers a unified perspective for assessing the effectiveness of various attack and defense strategies.

Looking ahead, several promising avenues exist for extending the model privacy framework. 
Firstly, a model engaging in the query-response-based interaction also risks revealing its parameters, hyper-parameters \citep{tramer2016stealing, wang2018stealing}, and even the training data \citep{fredrikson2014privacy, fredrikson2015model}. Recognizing this vulnerability, an extension of model privacy is to modify the evaluation criteria to include the protection of any model-related quantities. 
Secondly, instead of stealing the defender's model $\fd$, an attacker may be interested in a more ambitious goal of stealing the function $\ft$ underlying the training data based on which $\fd$ was obtained, such as demonstrated in our hate speech detection experiment. The model privacy framework can be adapted to this scenario by assessing the differences between the reconstructed function $\hat{f}_n$ and $\ft$, instead of $\fd$.
Thirdly, while this paper focuses on batch query strategies, designing optimal attacks and defenses under sequential query strategies remains an open yet crucial challenge.

% Driven by practical needs, we propose evaluation metrics to assess an attacker's cost of stealing a model. An attack is considered potent if the attacker can reverse-engineer the defender's model at a lower cost than gathering data and training a comparable model independently. 

\section*{Data Availability}
All real-world datasets used in this work are publicly available, while simulated datasets can be reproduced by our released code. 

\section*{Acknowledgments}
The work was supported in part by the National Science Foundation CAREER Program under grant number 2338506.

\section*{Conflict of Interest}
We declare that there is no conflict of interest.

% \section*{Submission Notes}
% abstract < 100 words, short title < 50 characters, cover letter, supplementary material \url{https://academic.oup.com/jrsssb/pages/general-instructions}

% \bibliographystyle{plain}
\bibliographystyle{abbrvnat}
\bibliography{privacy}

% \clearpage
% \listoffigures
\clearpage
\section*{List of Figure Legends}

\newcommand{\figurelegend}[2]{%
  \par\noindent\textbf{Figure #1.} #2\par\vspace{0.75\baselineskip}
}

\figurelegend{1}{Goodness comparison of different defense mechanisms against an attacker using polynomial regression.}
\figurelegend{2}{The unprotected responses and perturbed responses by proposed defense mechanisms.}
\figurelegend{3}{Goodness comparison of different defense mechanisms against an attacker performing penalized regression.}
\figurelegend{4}{Variable selection reliability of different defense mechanisms against an attacker performing penalized regression.}
\figurelegend{5}{Average accuracy of the attacker’s rebuilt
model for the hate speech detection task.}
\figurelegend{6}{Average F1 score of the attacker’s rebuilt model for the hate speech detection task.}

% Figure Legends
% 1 212 213 244 276 

%%%%%%%%%%%%%%%%%%%%%%%%%%%%%%%%%%%%%%%%%%%%%%%
% Appendix
%%%%%%%%%%%%%%%%%%%%%%%%%%%%%%%%%%%%%%%%%%%%%%%
% \newpage
% % \appendix  
% %     \renewcommand{\appendixname}{Appendix~\Alph{section}}  
% \begin{appendices}

% % \addcontentsline{toc}{subsection}{Supplement Contents}
% % \tableofcontents

% \input{supp.tex}
% \end{appendices}

%USE THE BELOW OPTIONS IN CASE YOU NEED AUTHOR YEAR FORMAT.
%\bibliographystyle{abbrvnat}
%\bibliography{reference}

%% sample for biography with author's image
% \begin{biography}{{\color{black!20}\rule{77pt}{77pt}}}{\author{Author Name.} This is sample author biography text. The values provided in the optional argument are meant for sample purposes. There is no need to include the width and height of an image in the optional argument for live articles. This is sample author biography text this is sample author biography text this is sample author biography text this is sample author biography text this is sample author biography text this is sample author biography text this is sample author biography text this is sample author biography text.}
% \end{biography}

% %% sample for biography without author's image
% \begin{biography}{}{\author{Author Name.} This is sample author biography text this is sample author biography text this is sample author biography text this is sample author biography text this is sample author biography text this is sample author biography text this is sample author biography text this is sample author biography text.}
% \end{biography}

\end{document}